\documentclass[a4paper,11pt]{article}
\usepackage[english]{babel}   
\usepackage[T1]{fontenc}
\usepackage{amsmath}
\usepackage{latexsym}
\usepackage{amssymb}
\usepackage{amsfonts}
\usepackage{subfigure}
\usepackage{graphics}
\usepackage{epsfig}

\usepackage{algorithm}
\usepackage{algpseudocode}
\usepackage{enumitem}
\usepackage{kmisc}

\def\w{{\mathbf w}}

\makeatletter
\def\hlinewd#1{%
\noalign{\ifnum0=`}\fi\hrule \@height #1 %
\futurelet\reserved@a\@xhline}
\makeatother

\begin{document}

%%%%%%%%%%%%%%%%%%%%%%%%%%%%%%%%%%%%%%%%%%%%%%%%%%%%%%%%%%%%%%%%%%%%%%%%%%%%%%%

\begin{flushright}
Submitted to IEEE Transactions on Image Processing on June 4, 2014
\end{flushright}

\begin{center}

\centerline{\rule[0mm]{16.5cm}{0.5mm}}
~\\
{\large {\bf  Aggregation of Local Parametric Candidates with Exemplar-based\\ Occlusion Handling for Optical Flow}} 
\footnote{This work was realized as part of the Quaero program, funded by OSEO, French State agency for innovation. It was also partly supported by Institut Curie, CNRS UMR 144 and the France-BioImaging project granted by the "Investissement d'Avenir" program.}\\~\\

{\sc Denis Fortun, Patrick Bouthemy and Charles Kervrann}\\
Inria, Centre Rennes - Bretagne Atlantique, Rennes, France\\~\\~\\
July 16, 2014\\

~\\
\begin{abstract}
Handling all together large displacements, motion details and occlusions remains an open issue for reliable computation of optical flow in a video sequence.
We propose a two-step aggregation paradigm to address this problem. The idea is to supply local motion candidates at every pixel  in a first step, and then
combine them to determine the global optical flow field in a second step. We exploit  local parametric estimations combined with patch correspondences
and we experimentally demonstrate that  they are sufficient to produce highly accurate motion candidates. The aggregation step is designed as the discrete optimization of a global regularized energy. The occlusion map is estimated jointly with the flow field throughout the two steps.
We propose a generic exemplar-based approach for occlusion filling with motion vectors.
We achieve state-of-the-art results in computer vision benchmarks, with particularly significant improvements in the case of large displacements and occlusions.\\~\\

\end{abstract}
~\\

\centerline{\rule[10mm]{16.5cm}{0.5mm}}
\end{center} 
 
%%%%%%%%%%%%%%%%%%%%%%%%%%%%%%%%%%%%%%%%%%%%%%%%%%%%%%%%%%%%%%%%%%%%%%%%%%%%%%%%%%%%%%%%%%%%%%%%%%%%%%%%%%%%%%%%%%%%%%%%%%%%%%%

\section{Introduction}
\label{sec:Intro}
Optical flow is a key information when addressing important problems in computer vision such as moving object segmentation, object tracking, egomotion
computation, obstacle detection or action recognition. The challenge for an optical flow estimation method is to deal with a large variety of image contents
and motion types. Optical flow has been historically evaluated on sequences exhibiting small displacements and smooth motion fields, like in the Yosemite
sequence \cite{Barron94}. Once initial issues were solved, other challenges were addressed \cite{Mitiche96}, and new situations have been proposed by new
benchmarks \cite{Baker11,Butler12}. Various and sometimes opposite motion conditions must be handled together, as illumination changes, large areas of smooth motion, motion details, large displacements, motion discontinuities, occluded regions (i.e., points disappearing in the next image).

Optical flow methods first rely on a data constancy assumption, e.g., applied to image intensity or spatial intensity gradient. Then, it is combined
with a spatial, or sometimes space-time, coherency constraint on the expected velocity field. Existing approaches can be broadly classified into \textit{local}
and \textit{global} methods.

Local spatial coherency arises when considering a parametric motion model, e.g., local translation \cite{Lucas81}, 4-parameter sub-affine model,
affine model, 8-parameter quadratic model \cite{Odobez95}, in a given neighborhood or an appropriate local region. Optimization
requires that the neighborhood is sufficiently textured or contains interest points such as corners, to supply accurate and reliable velocity vectors.

In contrast, global methods express the flow field coherency by imposing a global smoothness constraint in addition to the data constancy term, known as the
regularization term of the global energy as pioneered by \cite{Horn81,Nagel86}.
Global methods overcome uncertainty yielded by local supports in uniform intensity regions by diffusing motion from informative to non informative regions
via the global regularization constraint. The optimization problem of seminal model \cite{Horn81} was optimally solvable, but the estimation was
affected by oversmoothing and was limited to small displacements.

Numerous modifications of this original model, starting with \cite{Black93,Heitz93}, have been designed to resolve these two crucial issues, namely, handling
of large displacements and preservation of motion discontinuities. It was usually achieved by introducing a multi-resolution and incremental coarse-to-fine
framework along with piecewise smoothing or robust estimation. The data-driven term of the global optimization has also received attention. Image features like
image gradient \cite{Brox04}, texture component \cite{Wedel09} or Census transform \cite{Hafner13}, and matching criteria like Normalized Cross Correlation
(NCC) \cite{Werlberger10}, convey invariance properties to overcome limitations of the classical intensity constancy assumption. Several data-driven terms were
considered in \cite{Kim13} and managed by a locally adaptive fusion scheme. However, intricate optimization issues came with the increasing complexity of the modeling.
 
Existing local methods are far from being able to compete with global models in terms of accuracy in computer vision benchmarks. However, several works based
on joint estimation and segmentation of the motion field have shown that when appropriate segmented regions are found, affine models can be very accurate
representations \cite{Sun12,Unger12}. However, the alternate optimization schemes involved are sensitive to the initialization of the region supports.

In this paper, we define a new method for optical flow computation called \textit{AggregFlow} which exhibits several distinctive features.
First, we advocate the systematic computation of affine motion models over a set of size-varying square patches combined with patch-based pairings.
Indeed, we experimentally demonstrate that the sets of motion vectors computed that way comprise at least one accurate motion vector for each pixel.
On this basis, we build an optical flow estimation method composed of a first step computing local parametric candidates followed by a second step aggregating these candidates to produce the global flow field. The motion vector candidates are independently estimated on local supports without segmentation step. The aggregation is performed by a discrete optimization algorithm which selects one candidate at each pixel while ensuring piecewise smoothing of the resulting flow field.
 
Secondly, we address the occlusion problem in an original way by blending it with the motion estimation issue through the two steps of AggregFlow.
Motion candidates are extended in occluded areas with an exemplar-based technique. The estimated parametric model of the dominant motion in the image also contributes to create supplementary motion candidates. We extract local occlusion cues in the first step of AggregFlow and exploit them to guide the joint estimation of the occlusion map and motion field
in the aggregation step. Motion estimation in occluded regions is performed with a generic global exemplar-based approach. Specifically, we properly deal with
large displacements producing large occluded regions.

Our method can thus be viewed as a novel and efficient combination of local and global approaches for occlusion-aware optical flow computation.
The main original features and contributions of our method AggregFlow are listed below:
\begin{itemize}
\item Motion candidates are locally estimated by a general parametric patch-based method which ensures relevant and accurate motion vectors
at every point among all the computed candidates. 
\item Feature matching is integrated in an original and efficient way in the two-step aggregation framework.
\item We define a generic exemplar-based method for occlusion filling with motion vectors.
\item We propose a joint motion and occlusion estimation framework based on a sparse model guided by a local occlusion confidence map.
\item AggregFlow outperforms existing methods on the MPI Sintel benchmark which involves large displacements and occlusions, and it is competitive in the
Middlebury benchmark composed of videos depicting smaller movements.
\end{itemize}
A preliminary approach without any occlusion handling and dedicated to a specific application was presented in \cite{Fortun13}.

The paper is organized as follows. Section 2 describes related work. In Section~3, we present the parametric computation of motion candidates and the local
detection of occlusions. Section~4 is devoted to the aggregation stage. In Section~5, we report experimental results demonstrating the performance of AggregFlow. Section 6 contains concluding remarks.

%%%%%%%%%%%%%%%%%%%%%%%%%%%%%%%%%%%%%%%%%%%%%%%%%%%%%%%%%%%%%%%%%%%%%%%%%%%%%%%%%%%%%%%%%%%%%%%%%%%%%%%%%%%%%%%%%%%%%%%%%%%%%%%

\section{Related work}

Hereunder, we briefly review the literature on optical flow computation while focusing on issues related to our contributions.

\subsection{Feature correspondences and large displacements} 
The integration of feature correspondences in dense motion estimation has been investigated in several recent works. A first class of methods integrates
feature correspondences in a global energy model. Variational methods \cite{Braux13,Brox10,Heas08,Weinzaepfel13} include an additional term to a classical
global energy  to impose the flow to be close to pre-computed correspondences. Giving a fixed weight to the correspondences, this approach is sensitive
to matching errors. To overcome this problem, \cite{Braux13,Weinzaepfel13}	 focused on improving the matching step.
Another class of methods use correspondences to reduce the search space for discrete optimization and provide a coarse initialization for subsequent
refinement \cite{Chen13,Mozerov13,Xu12}. 
The main motivation of the attempts based on feature matching is to get rid of the drawbacks of the coarse-to-fine scheme imposed by variational optimization,
in particular the loss of large displacements of small objects.

Our patch correspondence is related to \cite{Chen13,Mozerov13,Xu12} in the sense that it is  used in the candidates generation process. However, our
candidates are not coarse approximations to be refined in a global subsequent step and we do not adopt any global variational
optimization.

\subsection{Occlusions}
Occlusions play a crucial role for motion estimation \cite{Stein09}, especially under large displacements, since no motion measurements are available
in occluded areas. Therefore, a proper occlusion handling must distinguish between \textit{occlusion detection}, segmenting the image into occluded
and non-occluded regions, and \textit{occlusion filling}, applying a specific treatment to motion estimation in occluded regions. Occlusion detection has been
mostly undertaken as a subsequent operation to motion computation, by thresholding a consistency measure issued from the estimated motion field, like geometric
forward-backward motion mismatch \cite{Ince08}, mapping unicity \cite{Xu12} or data constancy violation \cite{Xiao06}. Several flows and image criteria have been combined
in a learning framework \cite{Humayun11}. The main limitation of the latter is
that accuracy of occlusion detection is highly dependent on the quality of the initial motion estimation. To overcome this problem, other approaches estimate
the occlusion map jointly with the motion field \cite{Ayvaci12,Ince08,Kolmogorov01,Papadakis13}. Our occlusion detection falls in the latter category.

The problem of filling occluded regions with estimated velocity vectors when the occlusion map is known is closely related to the image inpainting problem.
Inpainting methods can be coarsely divided into two classes, diffusion-based methods \cite{Bertalmio00,Chan02} and exemplar-based methods \cite{Criminisi04,Komodakis07}. A synthesis of these two approaches has been investigated in \cite{Bugeau10} in a variational framework.
Occlusion filling is usually tackled by diffusion-based (or geometry-oriented) methods, propagating motion from non-occluded regions to occluded regions via partial derivative equation (PDE) resolution \cite{Ayvaci12,Ince08,Papadakis13,Xu12}. In exemplar-based image inpainting, the missing part is filled by copying pixels of the observed images. The framework is non local in the sense that similar pixels can be sought any where in the image. We adapt this strategy to occlusion filling with motion vectors.

\subsection{Parametric motion estimation}
The use of a parametric model has been widely investigated in motion estimation \cite{Black96,Cremers05,Fortun13,Memin98,Odobez95,Sun12}.
Applied on the whole image domain, affine or quadratic models are adequate to estimate the dominant image motion induced by the camera motion \cite{Odobez95}. For accurate dense motion estimation, parametric approximations are only valid locally. Local regions are usually defined as square patches centered on each pixel \cite{Black96,Lucas81}, possibly with an adaptation of the patch size \cite{Senst12}, or its position \cite{Jodoin09}. It has the merit of being easy to implement with a low computational cost,  but it is clearly outperformed by sophisticated extensions of \cite{Horn81} introduced in modern global optical flow methods.

As aforementioned, more complex region shapes can be estimated by joint motion segmentation and estimation. Existing approaches can be divided in two classes. A first class of methods relies on an independent image color segmentation and tries to fit parametric motion in each region
\cite{Black96b,Bleyer06,Gelgon00,Xu08,Zitnick05}, possibly with the help of an independent global variational estimation \cite{Black96b,Xu08}. The drawback
is that image color segmentation may lead to an over-segmentation of the motion field. The second class of methods jointly estimates region supports and
parametric motion models for each region \cite{Cremers05,Odobez98,Sun12,Unger12}. It is achieved by minimizing a global energy with respect to supports and motion
parameters of the regions. However, the global energy is highly non-convex and consequently difficult to minimize and particularly sensitive to the initialization of the optimization procedure.

The motion field produced by AggregFlow is composed of affine motion vectors estimated in square patches without any motion segmentation. AggregFlow
implicitly selects the best patch size and position when selecting the best motion candidate for each pixel in the second step.

\subsection{Motion discontinuities}
In the variational setting, the problem of preserving discontinuities has been addressed by modifying the regularization term \cite{Nagel86}. The seminal work
of \cite{Horn81} used a quadratic penalty function on the gradient magnitude of motion vectors. The first attempt to preserve discontinuities was investigated in
\cite{Heitz93} where a binary map of local motion discontinuities was introduced and estimated jointly with the motion field using two interwoven Markov Random
Fields (MRF). The regularization is thus canceled on motion discontinuities. Subsequent improvement has then been reached with the use of robust penalty
functions in the regularization term \cite{Black96,Memin98}. The robust $L_1$ norm is often retained in variational settings owing to its convexity
\cite{Brox04,Werlberger10}.

\subsection{Discrete optimization and aggregation paradigm}
Discrete optimization is an alternative to variational methods and is often able to find good local minima for non differentiable and
non-convex energy functionals. To combine the subpixel accuracy of the continuous variational approach and the efficiency of discrete minimization, the authors of
\cite{Lempitsky08} built a discrete motion space from motion fields delivered by several global variational estimations with different parameter settings.
An energy function is then optimized by successive fusions of global proposals, which are efficiently performed by a binary graph-cut method.
In \cite{Fortun12}, we followed a similar approach but with a semi-local patch-based variational estimation of candidate motion vectors. In \cite{Alba10}, a
set of candidate motion vectors is computed at each pixel using phase correlation in overlapping patches. The candidates are then linearly combined to create a
global motion field.
Recent works \cite{Chen13,Mozerov13} also exploit discrete graph-cut optimization in a two-step paradigm. However, the principle is
different than ours. Indeed, the motion candidate generation step only aims at finding dominant displacements and the aggregation provides a coarse initialization for a subsequent global refinement. Discrete optimization is also associated with a variational framework in \cite{Xu12} as an intermediate
stage between scales of a coarse-to-fine framework, in order to limit the loss of details of the flow.
Another aggregation-related work is the image colorization method of \cite{Bugeau14}. Color candidates are obtained with patch correspondences, and a candidate
is selected at each pixel by minimizing a global energy in a variational setting.

%%%%%%%%%%%%%%%%%%%%%%%%%%%%%%%%%%%%%%%%%%%%%%%%%%%%%%%%%%%%%%%%%%%%%%%%%%%%%%%%%%%%%%%%%%%%%%%%%%%%%%%%%%%%%%%%%%%%%%%%%%%%%%%

\section{Local motion candidates and occlusion cues}
\label{sec:candidates}

We describe in this section the first step of our method AggregFlow. It exploits local information to supply motion candidates and occlusion cues. 
A set of motion vector candidates is generated at every pixel by a combination of patch correspondences and local parametric motion model estimations. 
A specific treatment is applied to occluded regions by exemplar-based extension of the motion candidates set. We also exploit the dominant motion
in the image due to camera motion. Motion candidates and occlusion cues form the input of the second stage of AggregFlow described in Section~\ref{sec:Aggreg}.

Our approach can be viewed as a new way to address the problem of choosing the local neighborhood for parametric estimation. Rather than adapting the regions \textit{a priori} or jointly with the motion field, we operate in two steps: 1) estimation of motion candidates on several supports at every pixel, 2) implicit selection of the best support through the selection of the optimal candidate at each pixel within the aggregation step. 
In the sequel, we denote two consecutive image frames as $I_1,I_2:\Omega \rightarrow \mathbb R$, with $\Omega$ denoting the image domain.

\subsection{Local parametric motion candidates}
\label{sec:loc_param_est}
\subsubsection{Set of overlapping patches in $I_1$}
\label{sec:patch_distrib}
\begin{figure}[!t]
    \centering
    \includegraphics[width=10cm]{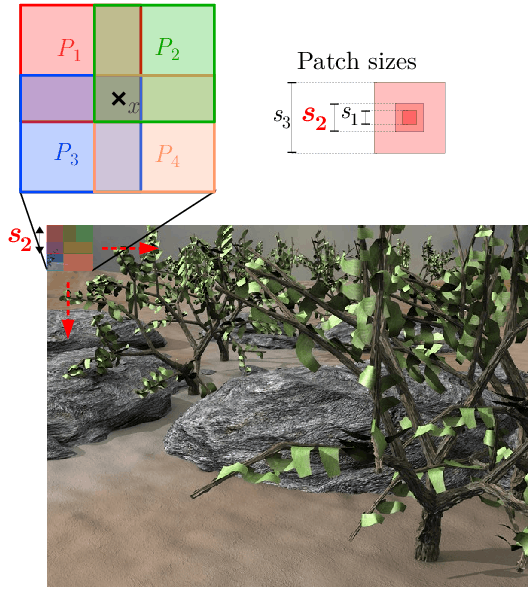}
    \caption{Four patches of set $\mathcal P_{s_2,\alpha}$ for a given size $s_2$ of the set $\mathcal S=\{s_1,s_2,s_3\}$, and overlapping ratio  
    $\alpha=0.3$. The pixel $x$ is contained in the patches $P_1,\ldots,P_4$. Motion estimation in each of these patches provide motion candidates for $x$.}
    \label{fig:patch_distrib}
    \end{figure}

The local supports for motion candidates computation are overlapping square patches of different sizes. Let us denote $\mathcal P_{s,\alpha}$ the patch set
for a fixed patch size $s$ and an overlapping ratio $\alpha \in [0,1]$ indicating the proportion of surface shared by neighboring patches (see illustration of
Fig. \ref{fig:patch_distrib}).
Let $\mathcal S = \{s_1 ,\ldots, s_n \}$ be a set of $n$ patch sizes, we then define $\mathcal P_{\mathcal S,\alpha} = \bigcup_{s\in \mathcal S} \mathcal
P_{s,\alpha}$.
To capture different motion scales, the patch sizes must cover a large range of values. In all our experiments, we will use $\mathcal S = \{16,44,104\}$.
Due to the overlap and the number of patch sizes ($n > 1$), one given pixel $x \in \Omega$ belongs to several patches. The motion vectors are estimated independently in each patch in two sub-steps described below: patch correspondences and affine motion estimations.\\

\subsubsection{Patch correspondences}
\label{sec:patch_corresp}
For each patch $P_1 \in \mathcal P_{\mathcal S,\alpha}$, we first determine the set $\mathcal M_N(P_1)$ of the $N$ most similar patches to $P_1$ in $I_2$.
Let us put forward that we do not aim at keeping at this stage the best correspondence only but at selecting $N$ relevant correspondences to subsequently constitute motion candidates. The matching step is generic and could be achieved with any arbitrary feature matching algorithm. 
We use a combination of the saturation and value channels of the HSV color space to  gain partial robustness to illumination changes \cite{Zimmer11} and we use the Sum of Absolute Distances (SAD) to compare patches. To avoid that the set $\mathcal M_N(P_1)$ uselessly contains too close patches, we impose a minimal distance between two patches of $\mathcal M_N(P_1)$. Hence, for each established pair of corresponding patches $P_{1,2} = (P_1 , P_2)$ with
$P_2 \in \mathcal M_N (P_1)$, we get the translation vector $\w_{P_{1,2}} \in \mathbb Z^2$ shifting $P_1$ onto $P_2$.

\subsubsection{Affine motion refinement}
\label{sec:patch_affine}
The displacements estimated by patch correspondences are integer-pixel translational approximations. To reach subpixel accuracy and to allow for more
complex motion, we refine the first sub-step of coarse translational motion vector $\w_{P_{1,2}}$ with the estimation of a local affine motion model
in every pair $P_{1,2}$. Denoting $\Omega_{P_1}$ the pixel domain of $P_1$, the affine motion model $\delta \w_{P_{1,2}} : \Omega_{P_1} \rightarrow
\mathbb R^2$ between $P_1$ and $P_2$ is defined at a pixel $x = (x_1 , x_2 )^\top$ as:
\begin{equation}
\delta \w_{P_{1,2}} (x) = (a_1 + a_2 x_1 + a_3 x_2 , a_4 + a_5 x_1 + a_6 x_2)^\top.
\end{equation}
The parameter vector ${\boldsymbol \theta}_{P_{1,2}}=(a_1, a_2, a_3, a_4, a_5, a_6)^\top$ of the affine model is estimated assuming brightness constancy:
\begin{equation}
\widehat {\boldsymbol \theta}_{P_{1,2}} = \argmin_{{\boldsymbol \theta}_{P_{1,2}}} \int_{\Omega_{P_1}} \phi(P_2(x+\w_{P_{1,2}}+\delta \w_{P_{1,2}}(x)) - P_1(x)) dx
\label{eq:estim_param}
\end{equation}
where the penalty function $\phi(\cdot)$ is chosen as the robust Tukey's function. The problem (\ref{eq:estim_param}) is solved with the publicly available
Motion2D software\footnote{http://www.irisa.fr/vista/Motion2D/} \cite{Odobez95}, which implements a multi-resolution incremental minimization scheme
involving an IRLS (Iteratively Reweighted Least Squares) technique.

\subsubsection{Final set of motion candidates}
The above described two-step estimation is repeated for every patch of $\mathcal P_{\mathcal S,\alpha}$ and generates a set of candidate motion
vectors $\mathcal C(x)$ at each pixel $x \in \Omega$ defined as follows:
\begin{eqnarray}
\label{eq:C}
\mathcal C(x) = \{\w_{P_{1,2}}(x) + \delta \w_{P_{1,2}}(x): P_1 \in \mathcal P_{\mathcal S,\alpha}(x), P_2 \in \mathcal M_N(P_1)	\}, 
\end{eqnarray}
where $\mathcal P_{\mathcal S,\alpha}(x) = \left\{P \in \mathcal P_{\mathcal S,\alpha} : x \in P \right\}$.

Let us make a few comments on the estimation scheme for computing motion candidates.
A coarse motion estimation followed by a refinement step has been investigated in several previous works \cite{Chen13,Leordeanu13,Mozerov13},
but it has always been dedicated to global motion fields. In our case, the refinement is local and adapted to each patch correspondence. 
Classical local motion estimation methods based on \cite{Lucas81} also rely on square patches, but assign the computed motion vector only to the center
point of each patch. On the opposite, parametric motion estimation in segmented regions as in \cite{Cremers05} apply to regions of arbitrary shape.
Our patch distribution can be considered as an intermediate level between these two extremes. Indeed, we use square patches as in \cite{Lucas81} and thus
avoid the complex segmentation step. However, we exploit the whole vector field issued from the affine model estimated in each patch. As a consequence, every pixel inherits several motion candidates from the affine motion estimations performed in patches of different positions and sizes which the given pixel belongs to. Finally, in contrast to several other methods using feature correspondences \cite{Brox10,Chen13,Weinzaepfel13}, we do not select one single patch
correspondence but we keep the $N$ best ones.

The interest of the local set of motion candidates supplied by AggregFlow is three-fold.
First, the correspondence sub-step enables to capture large displacements even for small patch sizes. Thus, it allows us to correctly deal with small structures undergoing large displacements in contrast to coarse-to-fine schemes.
Second, by considering a large variety of patches, we get rid of the predefined choice of the local neighborhood encountered in parametric motion estimation.
The selection of the proper patch via its corresponding motion candidate is transferred to the aggregation stage.
Third, introducing patches of several sizes enables to tackle motion of different scales.

\subsection{Motion candidates in occluded areas}
\label{sec:occ_candidates}
The generation of motion candidates described in Section \ref{sec:loc_param_est} does not differentiate between occluded and non-occluded pixels.
For a given pixel $x$, if all the patches of $\mathcal P_{\mathcal S,\alpha}(x)$ mainly contain occluded pixels, there is no chance to correctly estimate
a relevant motion candidate at $x$ in that way. Therefore, we compute motion candidates in occluded regions in a specific manner.

Let us define the occlusion map $o:\Omega \rightarrow \{0,1\}$ 
\begin{equation}
\label{occ}
o(x) = 
\begin{cases}
1 & \text{if} \; x \; \text{is occluded},\\
0 & \text{otherwise}.
\end{cases}
\end{equation}
The occluded regions are denoted $\mathcal O = \{x\in\Omega : o(x) = 1\}$. The computation of map $o$ will be addressed in Section \ref{sec::occ_conf} and
Section \ref{sec:Aggreg}, and we assume for now that $o$ is known.\\

\subsubsection{Occlusion filling with motion vectors}
When occluded regions are known, occlusion filling with motion vectors is conceptually closely related to image inpainting, since it recovers motion in regions where motion is by definition \textit{not observable}: The occluded pixels do not appear in the next image and consequently have no corresponding points. Classical methods for motion-based occlusion filling operate in a variational framework by cancelling the data term and letting the diffusion process of the regularization propagate the optical flow in occluded regions \cite{Ayvaci12,Xu12}. The diffusion-based class of inpainting methods \cite{Bertalmio00} acts similarly. They perform well in case of thin missing areas or cartoon-like images, but they are usually outperformed by exemplar-based inpainting methods \cite{Criminisi04} for large missing regions.
In order to deal with large occlusions produced by large displacements, we follow the inpainting analogy and we overcome the problem of local motion candidates estimation in occluded areas by designing an exemplar-based scheme. In the first step of AggregFlow, the motion candidates set is thus augmented by copy-paste operations.\\

\subsubsection{Exemplar-based candidates extension}
\label{sec:exemplar_candidates}

We rely on the assumption that motion at an occluded pixel $x \in \mathcal O$ is similar to the motion of a close non-occluded pixel $m_o(x) \in
\Omega\backslash\mathcal O$ belonging to the same object or the same background part. To provide relevant motion candidates at $x$, we copy motion
candidates from $\mathcal C(m_o(x))$ to $\mathcal C(x)$. The search domain ${\mathcal V_o} \subset \Omega\backslash\mathcal O$ for $m_o(x)$ is constrained
to be close to the occlusion boundaries. Figure \ref{fig:occ_exemplar1}(e) represents the occluded regions $\mathcal O$ (in white) and the search domain
${\mathcal V_o}$ (in red), and Fig. \ref{fig:occ_exemplar1}(f) superimposes the two sets on $I_1$.
Searching for pixel $m_o(x)$ for $x\in\mathcal O$ is actually easier for motion-based occlusion filling than for image inpainting. Indeed, occluded regions are not completely uninformative, while  inpainted regions are, since we have access to the information supplied by image $I_1$ even in $\mathcal O$. Thus, as $m_o(x)$ is expected to belong to the same object as $x$, we use color similarity to find the match in $I_1$:
\begin{equation}
\label{eq:match}
m_o(x) = \argmin_{y\in{\mathcal V_o}} D(I_1,x,y),
\end{equation}
where $D(I_1,x,y)$ is the distance between patches centered respectively in $x$ and $y$. As in Section \ref{sec:loc_param_est}, we resort to a SAD in
the HSV space.

An extended candidate set $\mathcal C_+(x)$ is created for occluded pixels by adding to the initial set $\mathcal C(x)$ the motion
candidates of their matched pixel $m_o(x)$:
\begin{equation}
\label{eq:C+}
{\mathcal C}_+(x) = \mathcal C(x) \cup \mathcal C(m_o(x)),~\forall x\in\mathcal O.\\
\end{equation}
By convention, $\forall x\in\Omega\backslash\mathcal O$, ${\mathcal C}_+(x) = \mathcal C(x)$.\\

\subsubsection{Occlusions due to camera motion}
A particular class of occluded (or disappearing) regions occurs at image borders in the case of large camera motion (Fig. \ref{fig:occ_camera}).
We cope with this issue by estimating the dominant image motion due to camera motion. To do so, we use again the robust parametric estimation
described in Section \ref{sec:loc_param_est}, but now, we apply it to the whole image \cite{Odobez95}, to retrieve the dominant motion.
We found in our experiments that the quadratic model was more adequate to accurately cope with large and sometimes complex camera motion.
The resulting parametric motion field $\w_{cam}:\Omega\rightarrow\mathbb R^2$ is added to the motion candidates, and we end up
with the final set of motion candidates $\mathcal C_{f}$:
\begin{equation}
\label{eq:Cf}
 {\mathcal C_{f}}(x) = \mathcal C_+(x) \cup \{\w_{cam}(x)\},~\forall x \in\Omega.
\end{equation}
The camera motion candidates are mostly useful for occluded pixels, but it can sometimes provide relevant motion candidates in unoccluded regions
of the background as well, so that we finally add it to all pixels in $\Omega$.

\subsection{Best candidate flow}
\label{sec:BCF}

\begin{figure}[!t]
  \centering
  $\begin{array}{c@{\hspace{5pt}}c}
 \includegraphics[width=220pt]{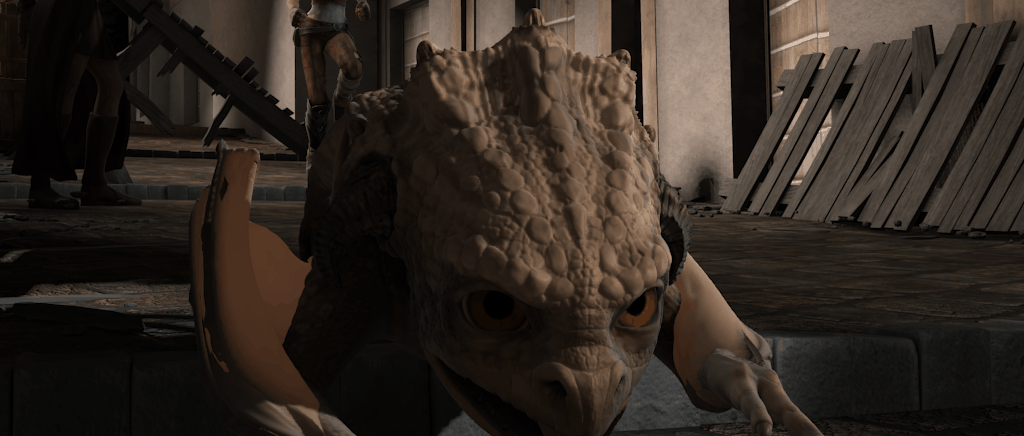} &
 \includegraphics[width=220pt]{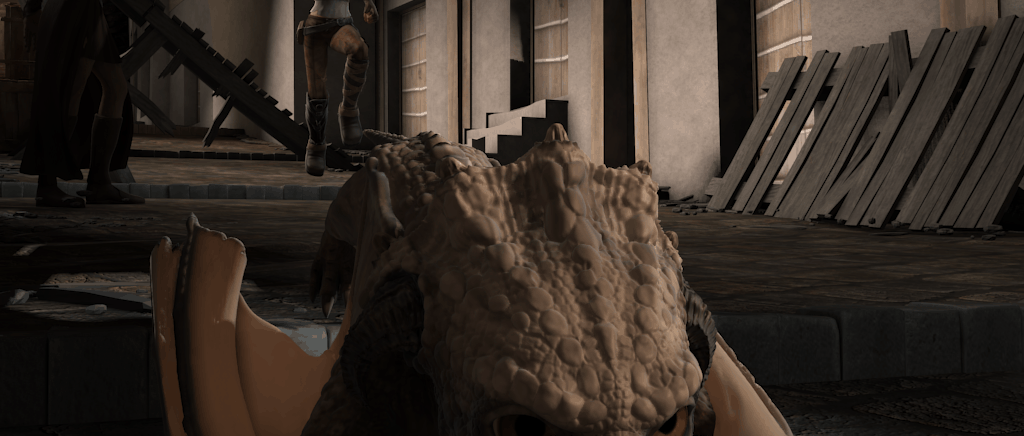} \\
\mbox{\small (a) $I_1$} & \mbox{\small (b) $I_2$} \\[5pt]
   \end{array}$
  $\begin{array}{c@{\hspace{5pt}}c}
 \includegraphics[width=220pt]{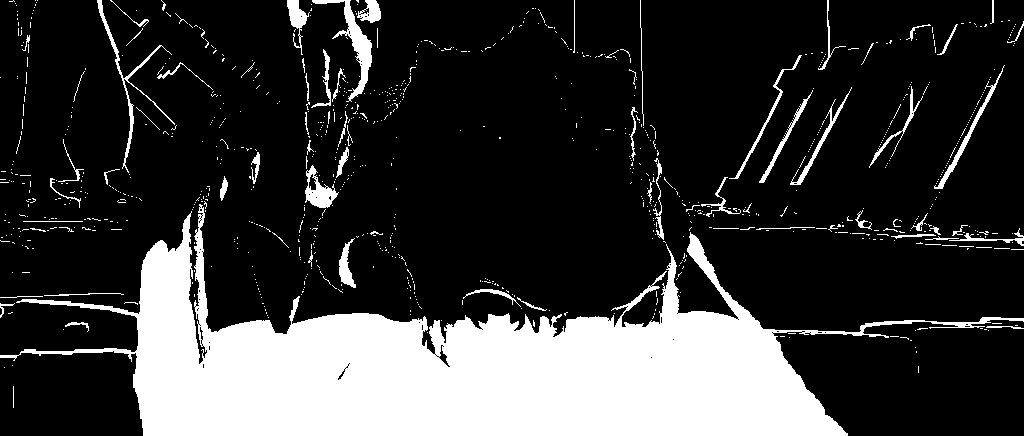} &
 \includegraphics[width=220pt]{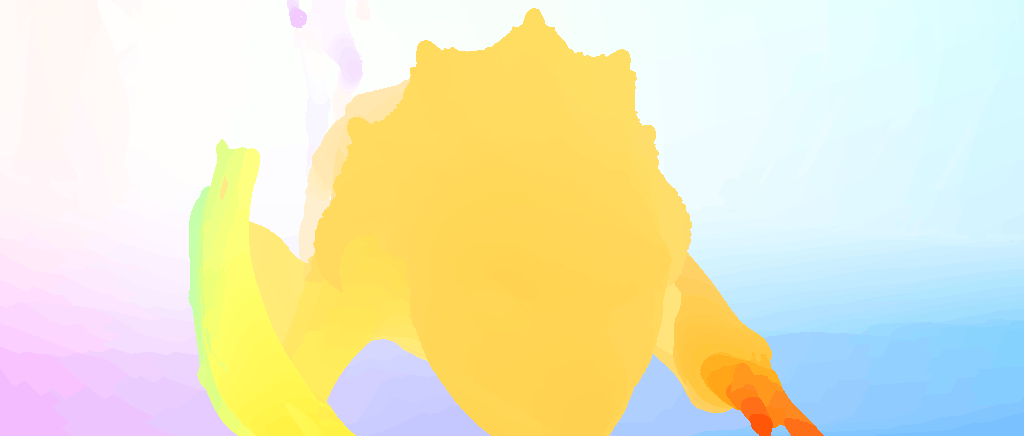} \\
 \mbox{\small (c) Ground-truth occlusion map} & \mbox{\small (d) Ground-truth $\w$} \\[5pt]
   \end{array}$
  $\begin{array}{c@{\hspace{5pt}}c}
 \includegraphics[width=220pt]{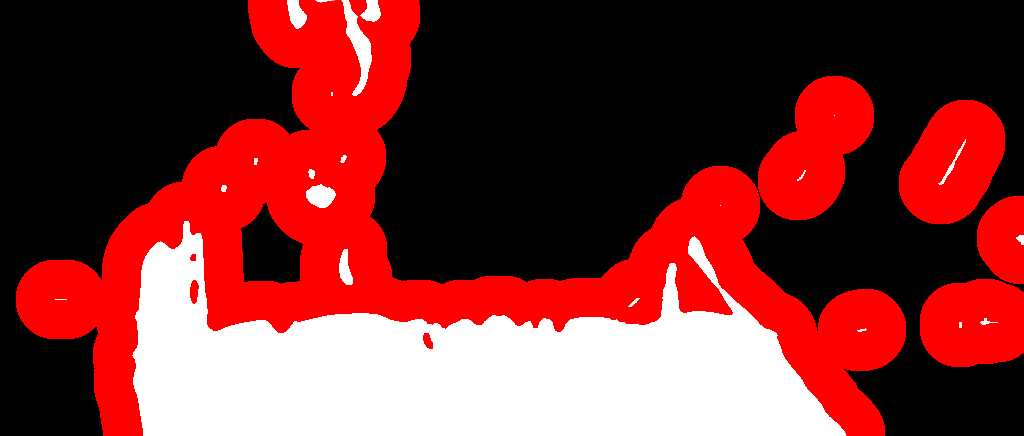} &
 \includegraphics[width=220pt]{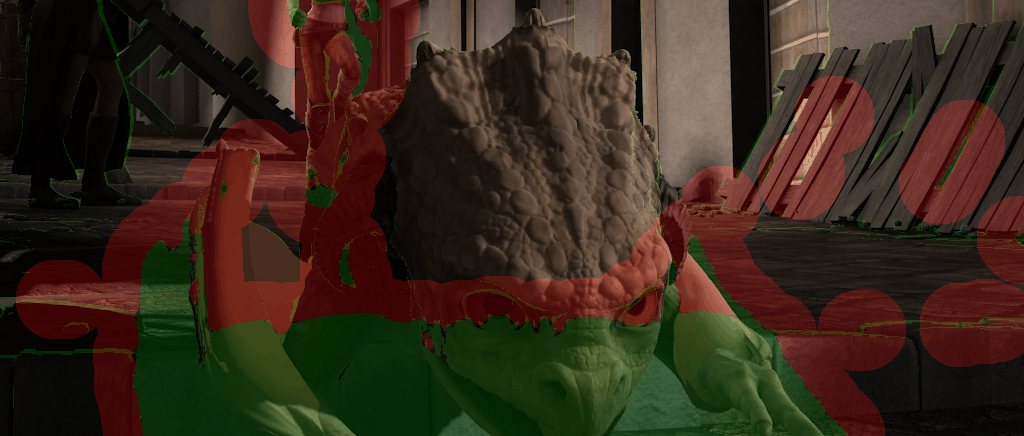} \\
 \mbox{\small (e) Occlusions (white)  } &  \mbox{\small (f) $I_1$ with occlusions (green) }  \\
 \mbox{\small  and search domain $\mathcal V_o$ (red)} & \mbox{\small  and search domain $\mathcal V_o$ (red)} \\[5pt]
   \end{array}$
  $\begin{array}{c@{\hspace{5pt}}c}
 \includegraphics[width=220pt]{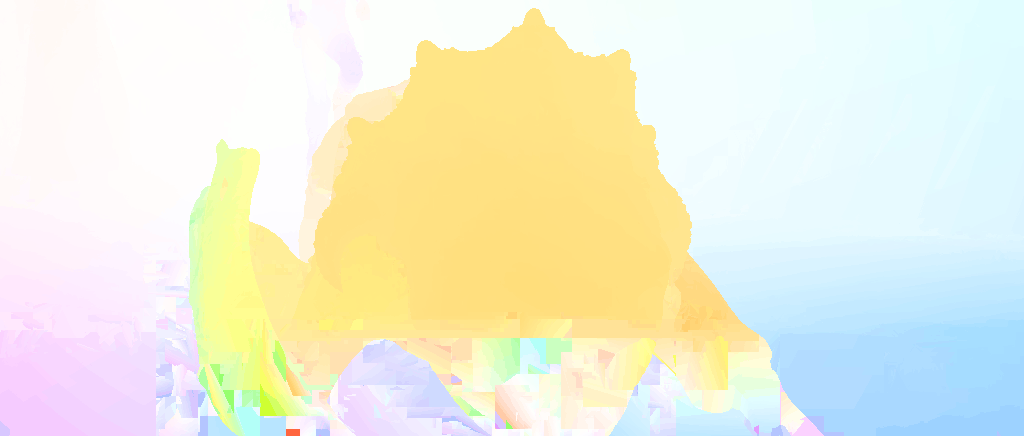} &
 \includegraphics[width=220pt]{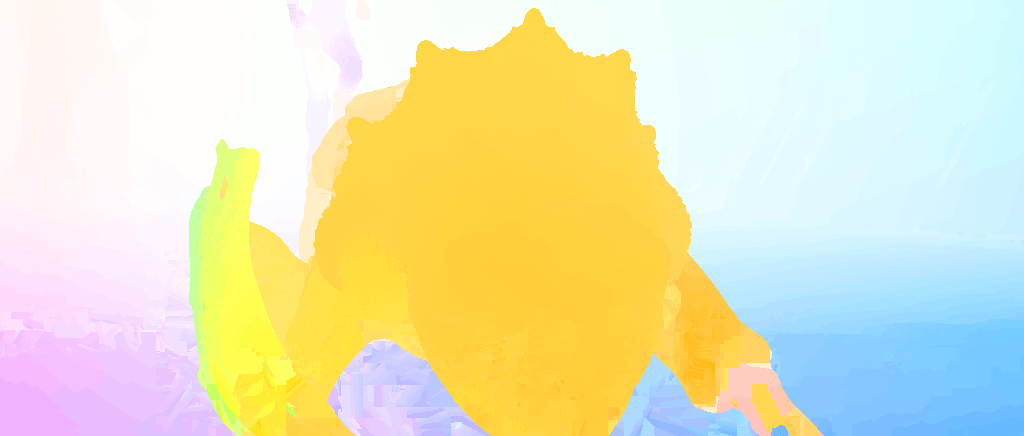} \\
  \mbox{\small (g) BCF without exemplar-based} & \mbox{\small (h) BCF with exemplar-based} \\
  \mbox{\small  candidates extension} & \mbox{\small  candidates extension} \\[10pt]
   \end{array}$

  \caption{Illustration of the performance improvement with exemplar-based candidates extension. First row: two successive input images. Second row:
  ground-truth occlusion map and motion field. Third row: representation of the search domain $\mathcal V_o$ (displayed here after median filtering of the occlusion map for the sake of visibility only). Fourth row: Best Candidate Flow obtained respectively without and with the exemplar-based candidates extension.}
\label{fig:occ_exemplar1}
  \end{figure}

 \begin{figure}[!t]
  \centering
   $\begin{array}{c@{\hspace{5pt}}c}
 \includegraphics[width=220pt]{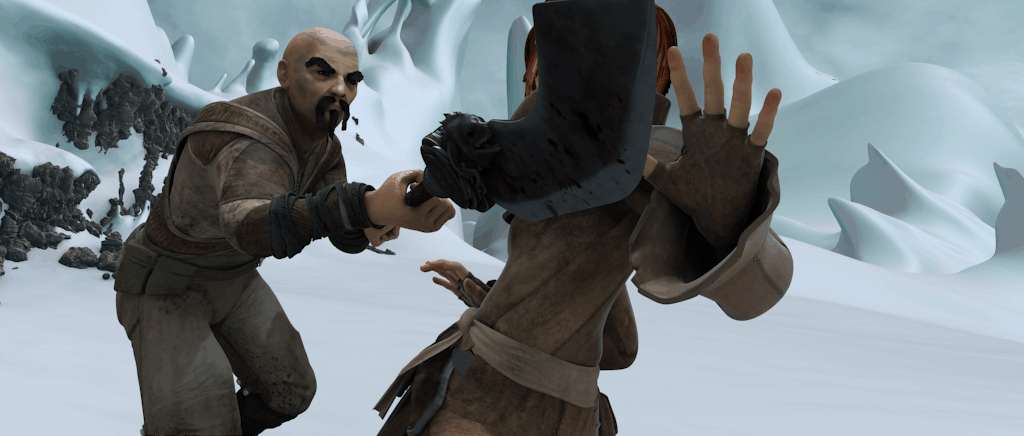} &
 \includegraphics[width=220pt]{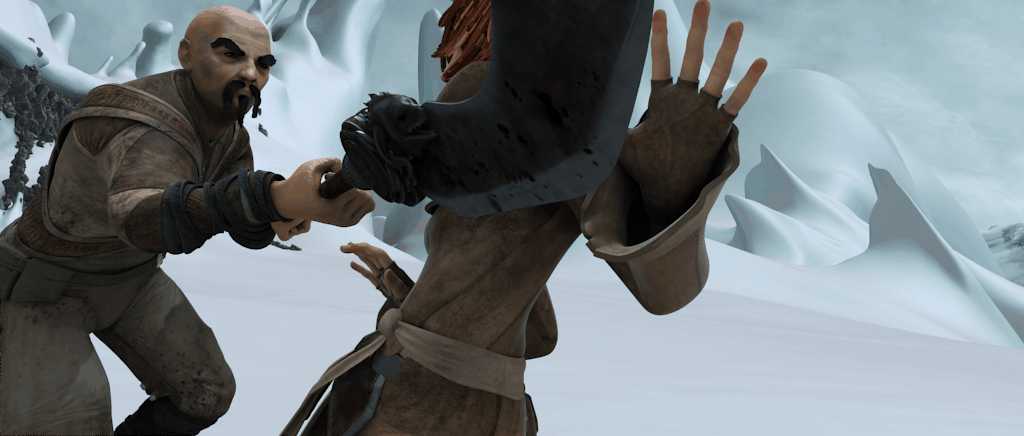} \\
\mbox{\small (a)  $I_1$} & \mbox{\small (b)  $I_2$}\\[5pt]
   \end{array}$
   $\begin{array}{c@{\hspace{5pt}}c}
 \includegraphics[width=220pt]{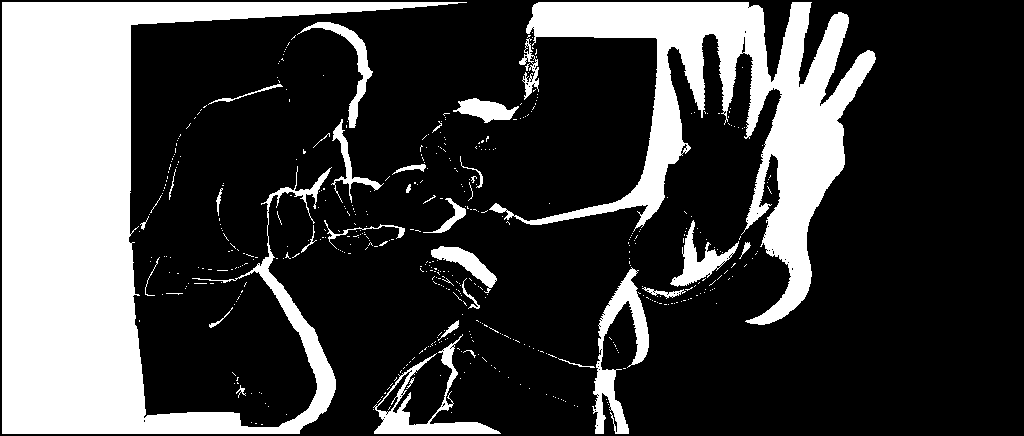} &
 \includegraphics[width=220pt]{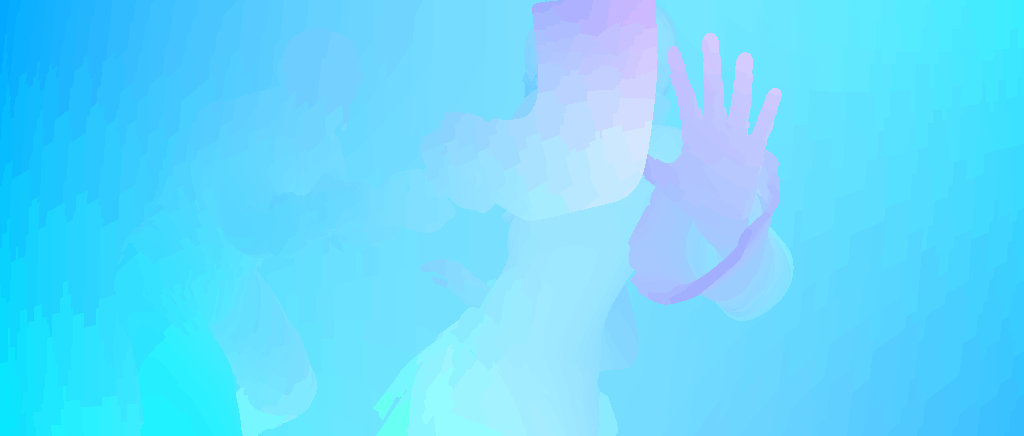} \\
 \mbox{\small (c)   Ground truth occlusion} & \mbox{\small (d)  Ground truth $\w$}\\[5pt]
   \end{array}$
	   $\begin{array}{c@{\hspace{5pt}}c}
 \includegraphics[width=220pt]{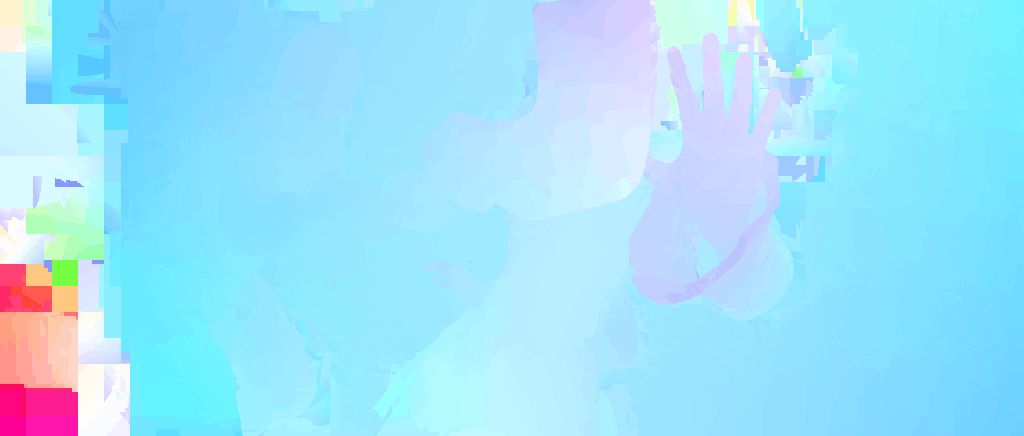} &
 \includegraphics[width=220pt]{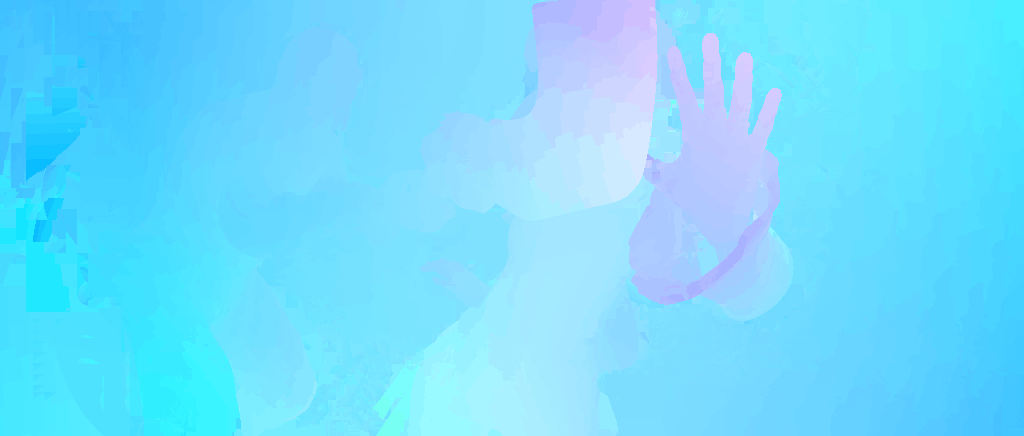} \\ [0cm]
  \mbox{\small (e)  BCF without camera motion} & \mbox{\small  (f) BCF with camera motion} \\
  \mbox{\small  candidates extension} & \mbox{\small  candidates extension} \\
   \end{array}$

  \caption{Performance improvement with camera motion candidates extension. First row: two successive input images. Second row: ground-truth occlusion map and motion field. Third row: Best Candidate Flow obtained respectively without and with the camera motion candidates extension.}
\label{fig:occ_camera}
  \end{figure}

To validate our method for computing motion candidates, we have exploited sequences from MPI Sintel and Middlebury datasets
\cite{Baker11,Butler12} provided with ground truth. We create the \textit{Best Candidate Flow} (BCF) by selecting at each pixel $x$ the candidate motion
vector of $\mathcal C_f(x)$ closest to the ground-truth vector. In order to evaluate our occlusion module, we distinguish between the BCF determined with the
candidates extension described in the preceding section (or full BCF) and the BCF without it. Parameters involved in the local motion computation are set to
$\mathcal S=\{16,44,104\}$, $\alpha = 0.75$, $N=2$.

Illustrations of the accuracy of the BCF are provided in Fig. \ref{fig:occ_exemplar1} and Fig. \ref{fig:occ_camera} on sequences of the MPI Sintel benchmark with large occluded regions. Besides, we make a specific focus on the improvements obtained with the candidates extensions. The difference between
BCF without candidates extension and the full BCF is clearly visible for occluded pixels and testifies the importance of the exemplar-based and camera motion
candidates extensions. Overall, the full BCF is very close to the ground-truth motion field revealing the performance of the local parametric motion
computation in the first step of AggregFlow. Indeed, we report in Table \ref{table:bcf} the objective evaluation given by the Endpoint Error (EPE) scores  for
the full BCF and BCF without candidates extensions, on the sequences provided with ground-truth in the datasets MPI {\sc Sintel} and {\sc Middlebury}. We also
compare them with those of motion fields supplied by \cite{Weinzaepfel13,Xu12}, as obtained with publicly available code.
Both BCFs outperform state-of-the-art methods \cite{Weinzaepfel13,Xu12} in the two benchmarks. Accuracy is further significantly improved with full BCF,
especially for the MPI {\sc Sintel} sequences where large displacements and wide occluded regions are present. It demonstrates that the combination of local
affine estimations in square patches with patch correspondences as described in Section \ref{sec:loc_param_est}, is quite relevant and sufficient to recover
very accurate motion fields. The challenge now is to select the best velocity vector among the motion candidates at every pixel.

\begin{table}[t]
 \centering
 \caption{EPE-all scores of motion fields on sequences with ground-truth from MPI MPI {\sc Sintel} and {\sc Middlebury} datasets.}
 \begin{tabular}{l|c|c}
 & MPI {\sc Sintel} & {\sc Middlebury} \\
\hlinewd{1pt}
 { {\bf Full BCF}} 						& {\bf 0.792} & {\bf 0.0710} \\
 { {\bf BCF w/o candidates extension}} 		& 1.851 & 0.0833 \\
  { DeepFlow \cite{Weinzaepfel13}} 	& 4.691 & 0.386 \\
 {  MDP-Flow2 \cite{Xu12}} 			& 4.006 & 0.223 \\
~\\
 \end{tabular}
  \label{table:bcf}
 \end{table}

\subsection{Occlusion confidence map}\,\, 
\label{sec::occ_conf}
In Section \ref{sec:occ_candidates}, the occlusion map $o$ was assumed to be known, and we addressed the motion-based occlusion filling problem
by recovering motion candidates for occluded pixels from non-occluded areas. Occlusion detection, that is the determination of $o$, will be performed
through the two steps of AggregFlow. In the first step, we compute a coarse occlusion confidence map, which will be used in the aggregation to guide the estimation. Our procedure is simple and exploits the patch distribution $\mathcal P_{\mathcal S,\alpha}$ and the correspondences used for motion candidates estimation. Nevertheless, from a more general point of view, the coarse occlusion confidence map could be designed differently, e.g.,
in the framework of \cite{Kervrann11}.

We first perform a coarse occlusion detection at the patch level. We consider the smallest patch size $s_1$ of the set $\mathcal S$ defined
in Section \ref{sec:loc_param_est} and detect the occluded patches of the set $\mathcal P_{s_1,\alpha}$. A common and simple occlusion detection
consists in checking the consistency of forward and backward estimated motion vectors \cite{Humayun11,Ince08,Mozerov13}. We apply the same principle
to patches of  $\mathcal P_{s_1,\alpha}$. Simplifying the notations of Section \ref{sec:loc_param_est} for the sake of readability, let us denote
$T^f_P$ the forward displacement between a patch $P\subset I_1$ and its matched patch $M_P\subset I_2$, and $T^b_P$ the backward displacement between $M_P$
and its matched patch in $I_1$. The forward-backward consistency criterion states that the patch $P$ is occluded if $\|T^f_P + T^b_P\| > \nu$,
where $\nu$ is a threshold. We then infer a patch-based occlusion map $o_P$ as follows:
\begin{equation}
\label{o_P}	
o_P(x) =
\begin{cases}
1 & \text{if } \exists P \in \mathcal P_{s_1,\alpha}(x) \; \text{such that} \; P \; \text{is occluded} \\
0 & \text{otherwise.}
\end{cases} 
\end{equation}

\begin{figure}[t]
  \centering
   $\begin{array}{c@{\hspace{5pt}}c}
 \includegraphics[width=220pt]{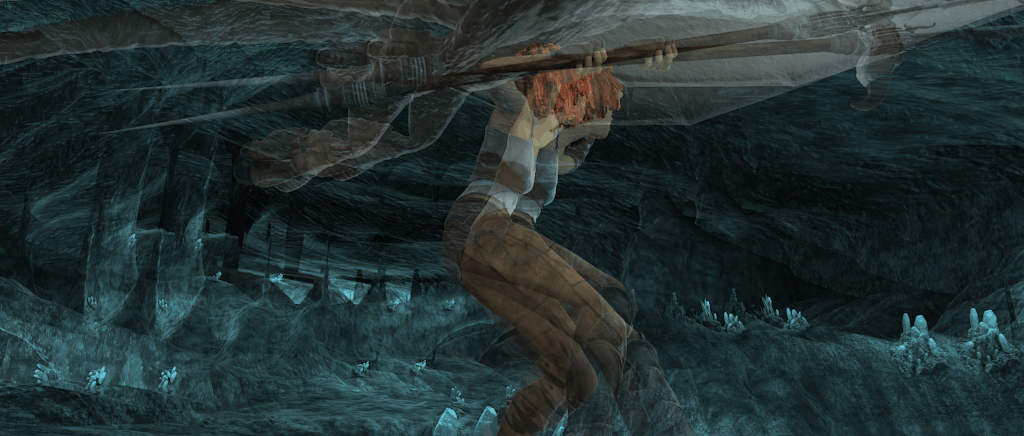} &
 \includegraphics[width=220pt]{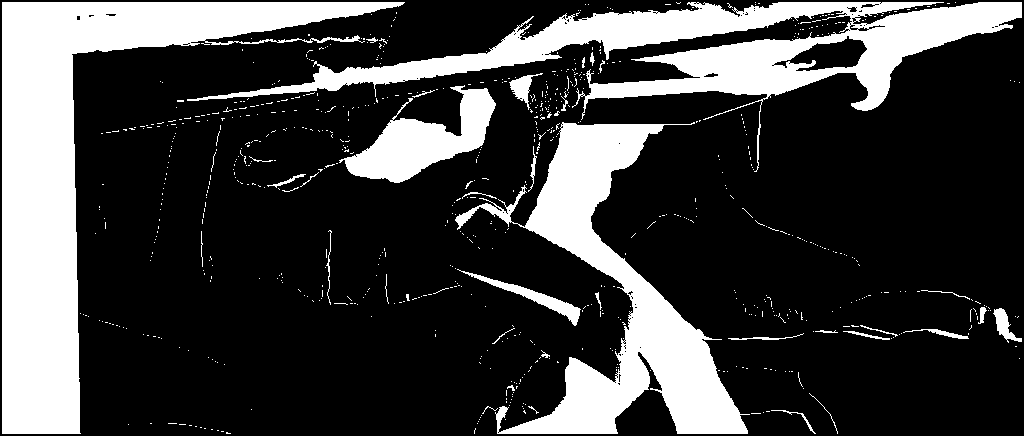} \\
\mbox{\small (a)  $I_1+I_2$} & \mbox{\small (b)  occlusion ground-truth}\\[5pt]
 \includegraphics[width=220pt]{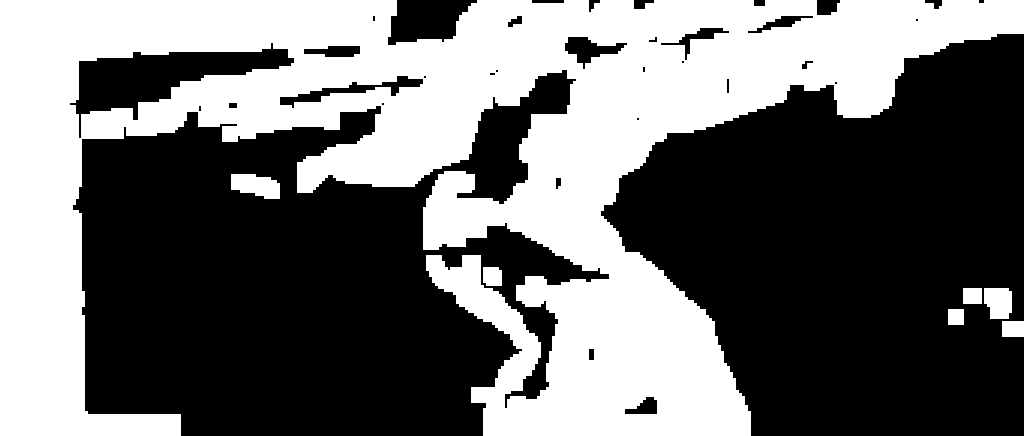} &
 \includegraphics[width=220pt]{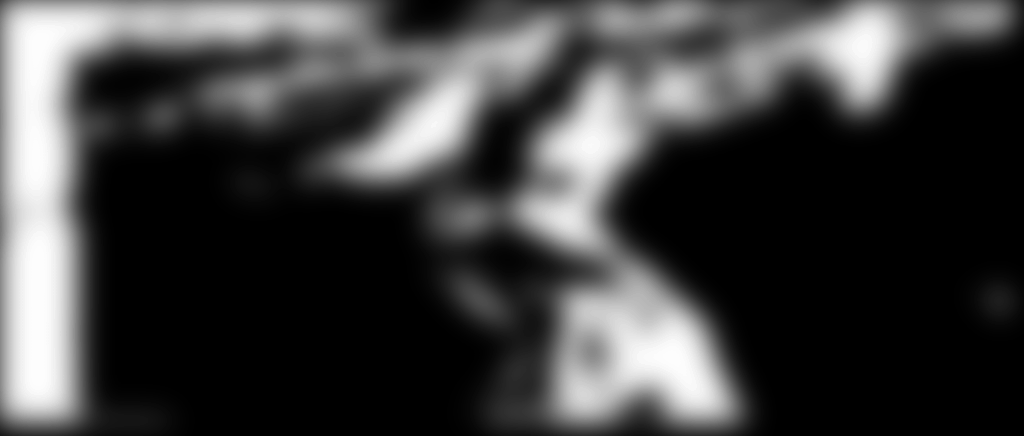} \\ [0cm]
\mbox{\small (c)  $o_P$ map } & \mbox{\small (d)  $\omega_o$ map} \\
   \end{array}$

  \caption{Patch-based occlusion detection. First row: Overlap of the two successive input images and occlusion ground-truth. Second row:
  Corresponding computed patch-based occlusion map $o_P$ and occlusion confidence map $\omega_o$.}
\label{fig:occ_conf}
  \end{figure}

Let us now consider the point set $\mathcal X_{o_P}$ composed of the centers of each occluded patch:
$\mathcal X_{o_P}=\{x\in\Omega:
o_P(x)=1, x\text{ is the center pixel of } P\}$.
We use the density of the point set as an indicator of the presence of occlusions.
We apply a Parzen density estimation on $\mathcal X_{o_P}=\{x_1,\ldots,x_{N_P}\}$, with $N_P$ the number of occluded patches:
\begin{equation}
\omega_o(x) = \frac{1}{N_P} \sum_{i=1}^{N_P} \frac{1}{\sigma}\; K\left(\frac{x - x_i}{\sigma}\right),
\label{eq:omega_o}
\end{equation}
where $\sigma$ is the bandwidth parameter and we choose $K$ to be a Gaussian kernel. We set $\sigma=s_1$. The occlusion confidence map $\omega_o$ is thus built as a probability density of the occlusion state. Figure \ref{fig:occ_conf} shows an example of $o_P$ and $\omega_o$. The map $\omega_o$ will be exploited in the aggregation stage to guide a sparsity-constrained occlusion detection.

The output of AggregFlow first step are the motion candidates set $\mathcal C_f(x)$ and the occlusion confidence map $\omega_o$. They will be exploited in the aggregation stage described in the next section to generate the final motion and occlusion fields.

%%%%%%%%%%%%%%%%%%%%%%%%%%%%%%%%%%%%%%%%%%%%%%%%%%%%%%%%%%%%%%%%%%%%%%%%%%%%%%%%%%%%%%%%%%%%%%%%%%%%%%%%%%%%%%%%%%%%%%%%%%%%%%%

\section{Discrete aggregation}
\label{sec:Aggreg}
The analysis of the Best Candidate Flow in subsection \ref{sec:BCF} has shown that the set of candidates at each pixel contains at least one motion vector
very close to the ground truth. Therefore, we conceive the aggregation as the selection of the best candidate at every pixel. To this end, we formulate
the aggregation as a discrete optimization problem, where the discrete finite motion vector space at each pixel $x$ is composed of the motion candidates
$\mathcal C_f(x)$. The occlusion map will be estimated jointly with the motion field while exploiting the occlusion confidence map $\omega_o$. The aggregation step amounts to minimizing the global energy function $E({\bf w},o)$:
\begin{eqnarray}
\label{aggreg}
\left	\{\widehat{\bf w},\widehat o\right\} = \argmin_{\{{\bf w},o\}}\; E({\bf w},o)~~
 \text{ s.t. } {\bf w}(x) \in \mathcal C_f(x)\text, \; o(x) \in \{0,1\}.
\nonumber
\end{eqnarray}
In the following, we detail the design of $E({\bf w},o)$ and the optimization strategy we have adopted.

\subsection{Global energy}
The aggregation energy is composed of four terms:
\begin{eqnarray}
E({\bf w},o) &=& E_{data}({\bf w},o,I_1,I_2) ~+~ E_{occ}(o,\omega_o) ~+~   E^{\w}_{reg}({\bf w}) ~+~ E^{o}_{reg}(o).
 \label{eq:Eaggreg}
\end{eqnarray}
\subsubsection{Data term $E_{data}$}
\label{sec:data_term}
The data term accounts for the relations between motion, occlusion and input images.
At non-occluded pixels, i.e.,  $o(x) = 0$, we rely on the usual constancy assumption of image intensity and of spatial image gradient, and we robustly penalize
the deviation from the data constraints.
The potential $\rho_{vis}$ associated to non-occluded (or visible) pixels is given by:
\begin{eqnarray}
\rho_{vis}(x,{\bf w}) = \phi(I_2(x+{\bf w}(x)) - I_1(x))	 ~+~ \gamma\phi(\nabla I_2(x+{\bf w}(x)) - \nabla I_1(x)),
\end{eqnarray}
where $\phi$ is the $L_1$ norm and $\gamma$ balances intensity and gradient constancy potentials. Resorting to discrete optimization allows us to use the
non-linearized brightness constancy equation. Thus, coarse-to-fine scheme is not required to cope with large displacements, and we avoid drawbacks related to the loss of small objects with large displacements.

At occluded pixels, no correspondence can be established by definition, and consequently none image feature constancy constraint can be exploited.  Therefore, coherently with the motion candidate extension of the first step, we define an exemplar-based data term for occluded pixels, encoded in the potential
$\rho_{occ}$:
\begin{equation}
\label{rho_occ}
\rho_{occ}(x,{\bf w},m) ~=~ \left\|{\bf w}(x) - {\bf w}(m(x))\right\|^2,
\end{equation}
where $m(x)$ is the visible pixel matched with pixel $x$ as obtained in (\ref{eq:match}). The motion vector of an occluded pixel is thus expected to be
similar to the motion vector of its matched non-occluded pixel.
The data term is finally formed by incorporating the selection of either the visible or the occlusion potential using the occlusion map:
\begin{eqnarray}
\label{eq:Edata}
E_{data}({\bf {\bf w}},o,I_1,I_2) = \sum_{x\in\Omega}(1-o(x))\,\rho_{vis}(x,{\bf w}) ~+~ \lambda_1 \, o(x)\, \rho_{occ}(x,{\bf w},m).
\end{eqnarray}
In contrast to other occlusion filling methods which only cancel the visibility term $\rho_{vis}$ in occluded areas and fill the occlusions with motion vectors
by diffusion \cite{Ayvaci12,Papadakis13,Xu12}, $\rho_{occ}$ acts as a valid data term at occluded pixels.

Concerning the occlusion recovery (i.e., the optimization w.r.t. $o$), the data term favors the selection of the occluded label at pixels where the data constancy
term is strongly violated. The continuous approach of \cite{Ayvaci12} operates in a similar way. In \cite{Ayvaci12}, the data constancy deviation
is balanced by an estimated continuous residual intensity field, from which occluded points are retrieved by thresholding. In contrast, our occlusion map is
binary by nature, and strongly prevents the influence of irrelevant data-constancy constraints on motion estimation in occluded areas.

\subsubsection{Occlusion term $E_{occ}$}
\label{sec:occ_constraint}
The data term (\ref{eq:Edata}) favours the detection of occluded pixels and must be counterbalanced by another term penalizing occlusion occurrence
defined by:
\begin{equation}
\label{eq:occ_constraint}
E_{occ}(o,\omega_o) = \lambda_2 \sum_x \omega_o(x) o(x),
\end{equation}
where $\omega_o$ is the occlusion confidence map computed in the first stage. The penalty of occlusion occurrence can be interpreted as a sparsity constraint
on the binary occlusion field $o$.
A sparsity constraint for occlusion detection was also proposed in \cite{Ayvaci12} in a continuous setting, and in \cite{Papadakis13} for a binary occlusion
variable, but without confidence map.

If we set $\forall x\in\Omega, \omega_o(x)=1$, which would be similar to what is done in \cite{Ayvaci12,Papadakis13}, the data-driven occlusion detection would
boil down to the data term (\ref{eq:Edata}) and (\ref{eq:occ_constraint}) would be a pure sparse prior constraint. The detection of the occlusion map would be
then too tightly coupled with the currently estimated motion field. We would face a chicken-and-egg problem, where $o$ is determined by $\w$, which also
depends on $o$. The consequence on the alternate optimization scheme would be a rapid trap into a local minimum.

\begin{figure}[!t]
  \centering
   $\begin{array}{c@{\hspace{5pt}}c}
 \includegraphics[width=175pt]{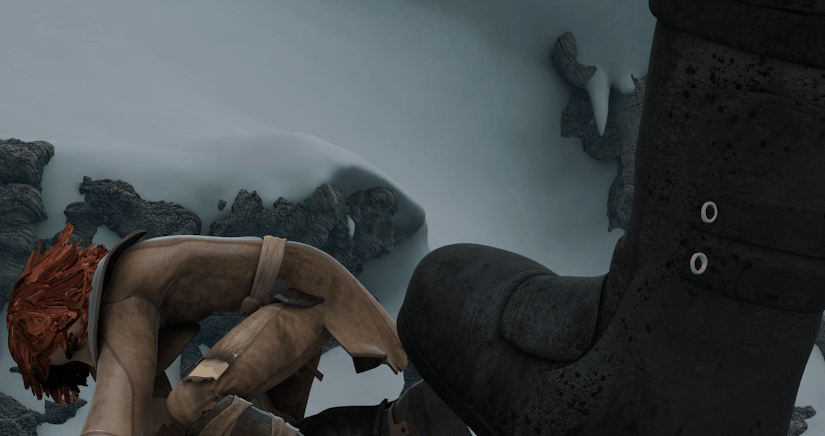} &
 \includegraphics[width=175pt]{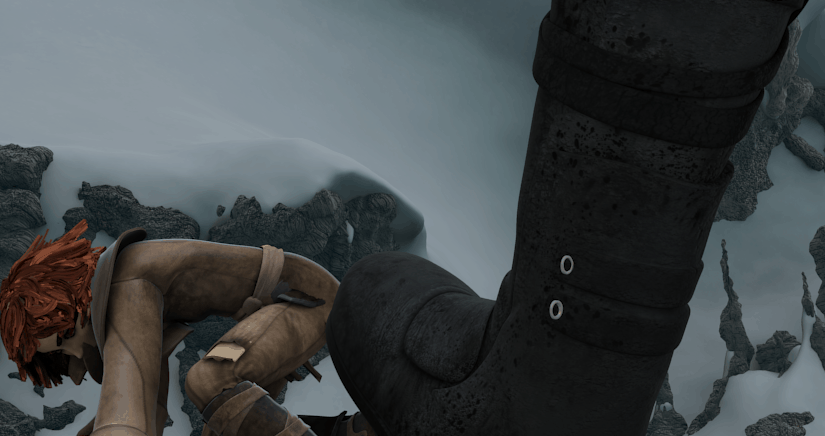} \\
\mbox{\small \small  (a) $I_1$} & \mbox{\small \small  (b) $I_2$}\\
 \includegraphics[width=175pt]{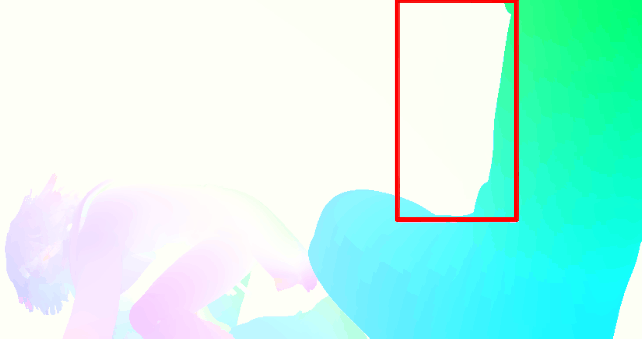} &
 \includegraphics[width=175pt]{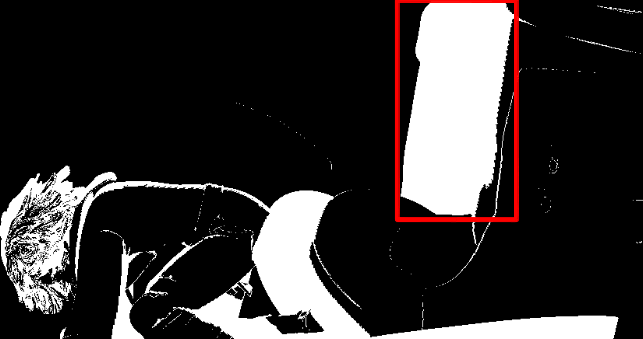} \\ [0cm]
\mbox{\small \small  (c) ground truth $\w$} & \mbox{\small \small   (d) Ground truth $o$} \\
 \includegraphics[width=175pt]{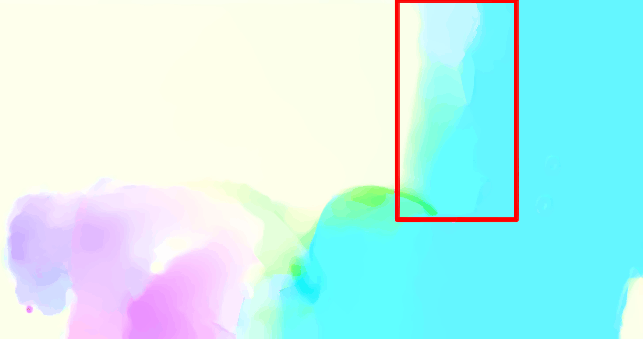} &
 \includegraphics[width=175pt]{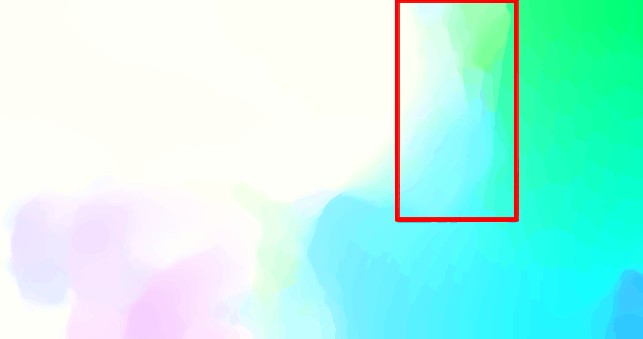} \\ [0cm]
\mbox{\small \small  (e) LDOF \cite{Brox10}} & \mbox{\small \small  (f) DeepFlow \cite{Weinzaepfel13}} \\
 \includegraphics[width=175pt]{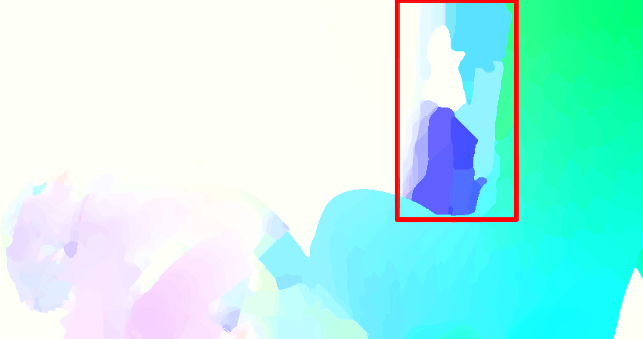} &
 \includegraphics[width=175pt]{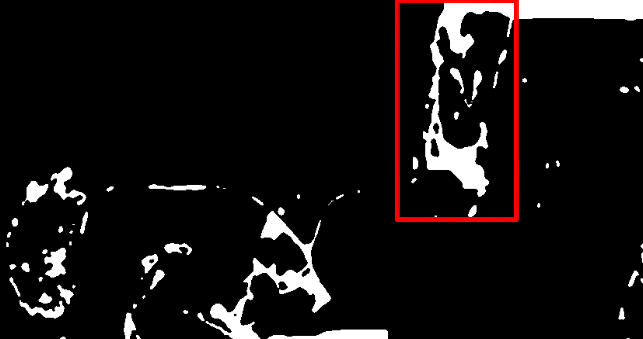}\\
\mbox{\small \small  (g) AggregFlow $\w$, without $\omega_o$} & \mbox{\small \small  (h) AggregFlow $o$, without $\omega_o$} \\
 \includegraphics[width=175pt]{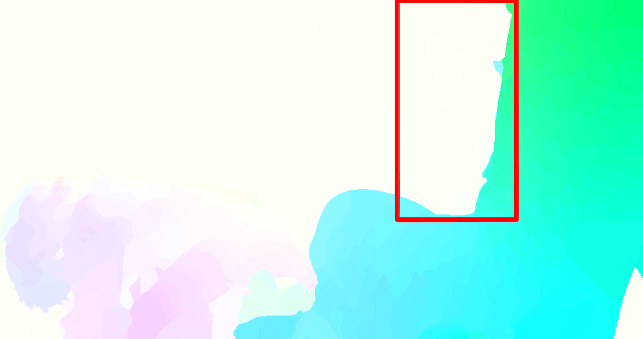} &
 \includegraphics[width=175pt]{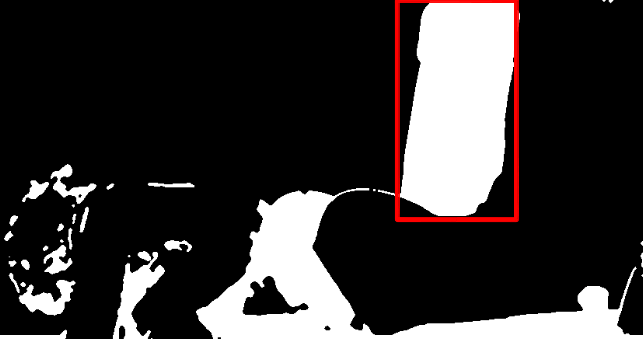} \\ [0cm]
\mbox{\small \small  (i) AggregFlow $\w$, with $\omega_o$} & \mbox{\small \small  (j) AggregFlow $o$, with $\omega_o$} \\
   \end{array}$

  \caption{Influence of the occlusion confidence map $\omega_o$ on motion and occlusion estimation. (e),(f): results of variational methods \cite{Brox10,Weinzaepfel13} without occlusion handling. (g),(h): similar behaviour of our method without occlusion confidence map and impact on the occlusion detection. (i),(j): output of AggregFlow when integrating the occlusion confidence map.}
\label{fig:occ_conf_final}
  \end{figure}

Illustrations are given in Fig. \ref{fig:occ_conf_final}. The results of two variational methods without occlusion handling \cite{Brox10,Weinzaepfel13} are displayed in Fig. \ref{fig:occ_conf_final} (e,f). In both cases, the motion field in the occluded region, highlighted by the red bounding box, is wrongly estimated
because no occlusion detection is performed. If the occlusion map is initialized to $o(x)=0, ~\forall x \in \Omega$, the occlusion terms of our energy (\ref{eq:Eaggreg}) are canceled in the very first iteration of the alternate optimization, which results in a similar behaviour to the methods \cite{Brox10,Weinzaepfel13}. If $\forall x\in\Omega, \omega_o(x)=1$, the convergence remains trapped in the initial local minimum, as displayed in Fig. \ref{fig:occ_conf_final} (g,h). The reason is that the occlusion map is determined by the motion field and cannot deviate from the output of the first iteration.
The role of the confidence map $\omega_o$ is then to act as an additional evidence for occlusion detection, relaxing the coupling between $\w$ an $o$. 
The guidance of $\omega_o$ enables to deviate from the output of the first iteration and to converge to the result shown in Fig. \ref{fig:occ_conf_final} (i,j).\\

\subsubsection{Regularization terms $E^{\w}_{reg}$ and $E^o_{reg}$}
The term $E^{\w}_{reg}({\bf  w})$ enforces piecewise smoothness of the motion field:
\begin{equation}
\label{Ereg}
E^{\w}_{reg}({\bf w}) = \lambda_3 \sum_{<x,y>} \beta(x) \phi(\|{\bf w}(x)-{\bf w}(y)\|^2)
\end{equation}
where $<x,y>$ denotes the two-site clique issued from the 8-neighborhood system. The weights $\beta(x)$ are given by
$\beta(x) = \exp\left(-\|\nabla I^0_1(x)\|^2/\tau^2\right)$ to modulate the regularization according to the intensity edge strength. To limit the
influence of noise and textured regions on the weights, we consider a smoothed version $I^0_1$ of $I_1$ obtained with the $L_0$ smoothing of \cite{Xu11}, favouring piecewise constant images and preserving only the abrupt edges.

It is also important to impose smoothness of the occlusion map with the term $E_{reg}^o$:
\begin{equation}
E^o_{reg}(o) = \lambda_4 \sum_{<x,y>} (1 - \delta(o(x)=o(y))),
\end{equation}
where $\delta$ designates the Kronecker function equal to 1 if its argument is true. The term $E^o_{reg}(o)$ completes the exemplar-based occlusion filling
described in Section \ref{sec:data_term} with diffusion-based occlusion filling.

\subsection{Optimization} 
\label{sec:optim}
The optimization problem (\ref{aggreg}) is solved by alternating minimization w.r.t. ${\bf w}$ and $o$. The initial value of $o$ is given by the coarse patch-based occlusion detection $o_P$ defined in (\ref{o_P}). The matching variable $m$ attached to the exemplar-based candidates extension is initialized with
$m_o$ defined in (\ref{eq:match}) and is recomputed after each update of the occlusion map. Convergence was empirically observed after three iterations in most cases. To avoid unnecessary computational cost, we fix the number of iterations to 3 for all sequences. Table \ref{table:overview} gives an overview of AggregFlow. Hereafter, we give details on the minimization procedure concerning $\w$ and $o$.

Once $\widehat{\bf w}$ is fixed, the energy to optimize w.r.t. $o$ amounts to:
\begin{eqnarray}
\label{eq:min_o}
\min_o~ \sum_{x\in\Omega}(1-o(x))\,\rho_{vis}(x,\widehat{\bf w})  ~+~ \lambda_1 \, o(x)\, \rho_{occ}(x,\widehat{\bf w},m) ~~~~~~~~~~~~~~~~~~~~~~~~~~~~~~~~~\nonumber\\
~~~~~~~~~~~~~~~~~~~~~~~~~~~~~~+~ \lambda_2 \sum_x \omega_o(x) o(x) ~+~ \lambda_4 \sum_{<x,y>} (1 - \delta(o(x)=o(y))).
\end{eqnarray}
Since
the pairwise term is submodular, the problem (\ref{eq:min_o}) can be solved exactly with standard graph cut method \cite{Boykov01}.

The optimization w.r.t. $\w$ with $\widehat o$ fixed is more difficult. The reduced energy function writes:
\begin{eqnarray}
\label{eq:min_w}
\min_{\bf w}~ \sum_{x\in\Omega}(1-\widehat o(x))\,\rho_{vis}(x,{\bf w}) 
 ~+~ \lambda_1 \, \widehat o(x)\, \rho_{occ}(x,{\bf w},m) ~~~~~~~\;~~~~~~~~\nonumber\\
 ~+~ \lambda_3 \sum_{<x,y>} \beta(x) \phi(\|{\bf w}(x)-{\bf w}(y)\|^2).
\end{eqnarray}
The motion label space has the specificity to be huge (the size of the candidates set $\mathcal C_f(x)$ can exceeds 200), and to be spatially varying
(each set of motion candidates is specific to each pixel). Message passing methods like belief propagation \cite{Felzenszwalb06} and TRW-S \cite{Kolmogorov06} can be applied to spatially varying label sets, as investigated in \cite{Ulen13} for stereo, but we found these methods to be too slow for our minimization problem (\ref{eq:min_w}). An alternative is to resort to graph-cut move-making methods \cite{Boykov01}, generalized in \cite{Lempitsky08} to spatially varying label sets. In this setting, each \textit{move} is a binary optimization problem  defined on an auxiliary variable selecting between two global proposals. Due to the spatial variability of the proposals and their independence, the submodularity of the regularization potential of (\ref{eq:min_w}) cannot be ensured, and only suboptimal moves can be achieved using QPBO \cite{Rother07}. 

Our aggregation problem differs from the one of \cite{Lempitsky08} since our motion candidates are locally determined. In contrast, \cite{Lempitsky08} exploits
global flow fields that can be directly used as proposals in the move-making process. Thus, we have to build global flow field proposals at each iteration from
the local motion candidates computed in patches. The important point is to ensure to some degree spatial smoothness of the proposal.
To this end, at each iteration, we choose a size $s_i$ and we take a subset of $\mathcal P_{ s_i,\alpha}$ formed by non-overlapping patches. Then, we retain for every pixel $x$ the motion candidate from $\mathcal C_f(x)$ corresponding to the patch where pixel $x$ lies.
We build as many global proposals as necessary to explore the motion candidate space.

Another issue arises from the non-local interaction involved in the exemplar-based term $\rho_{occ}(x,{\bf w},m)$. To make the optimization problem
tractable, we transform $\rho_{occ}(x,{\bf w},m)$ to a pixel-wise term at each \textit{move-making} iteration by fixing the exemplar-based constraint
$\w(m(x))$ to its value at the previous iteration. At a given \textit{move-making} iteration $i$, denoting $\widehat\w^{(i-1)}$ the value of ${\bf w}$ at
iteration $i-1$, the potential becomes:
\begin{equation}
\rho_{occ}(x,{\bf w},m) = \left\|{\bf w}(x) - \widehat{\bf w}^{(i-1)}(m(x))\right\|^2.
\end{equation}

In the next section, we analyse the performance of AggregFlow
with experiments on challenging image sequences.

%%%%%%%%%%%%%%%%%%%%%%%%%%%%%%%%%%%%%%%%%%%%%%%%%%%%%%%%%%%%%%%%%%%%%%%%%%%%%%%%%%%%%%%%%%%%%%%%%%%%%%%%%%%%%%%%%%%%%%%%%%%%%%%

\section{Experimental results}

\begin{table}{t}
\label{table:overview}
\caption{Overview of AggregFlow.}
\vspace{0.15cm}
  \begin{center}
\fbox{
{
  \begin{minipage}{360pt}
\vspace{0.15cm}
  \begin{enumerate}[label=\arabic*., font=\bf]
    \item {\bf Local step}
    \begin{enumerate}[label*=\arabic*., font=\bf]
      \item Generate the motion candidates sets $\mathcal C(x)$ (\ref{eq:C})
      \item Compute patch-based occlusion map $o_P$ (\ref{o_P})\\
      Derive the occlusion confidence map $\omega_o$ from $o_P$ (\ref{eq:omega_o}) 
      \item Compute the matching variables $m_o(x)$ (\ref{eq:match})\\
      Extend motion candidates in occluded regions to obtain $\mathcal C_f($ (\ref{eq:Cf})
    \end{enumerate}
    \fbox{{Output of the $1^{st}$ step}: $\mathcal C_f$, $\omega_o$}
\vspace{0.1cm}
  \item {\bf Global aggregation}\\ Initialize $o=o_P$ and $m=m_{o}$\\ Iterate:
    \begin{enumerate}[label*=\arabic*., font=\bf]

      \item Estimate $\w$ (\ref{eq:min_w})
      \item Estimate $o$ (\ref{eq:min_o})
      \item Update $m$ (\ref{eq:match})
    \end{enumerate}
    \fbox{Output of the $2^{nd}$ step: $\w,o$}
\vspace{0.15cm}
  \item {\bf Post-processing} : weighted median filtering on $\w$\\
  \end{enumerate}

  \end{minipage}}
}
  \end{center}
\end{table}

\subsection{Implementation details}
All the patch correspondences involved in AggregFlow are computed with the PatchMatch algorithm \cite{Barnes09} based on the minimal C++ code provided by the authors\footnote{http://gfx.cs.princeton.edu/pubs/Barnes\_2009\_PAR/index.php}. A weighted median filtering with bilateral weights \cite{Xu12b} is performed as a post-processing step to enhance motion edges as advocated in \cite{Sun14}. For the discrete minimization, we use available QPBO and max-flow code\footnote{http://pub.ist.ac.at/~vnk/software.html}. After extensive experimental tests, the aggregation parameters have been set to $\lambda_1=5$,
$\lambda_2=50$, $\lambda_3=500$, $\lambda_4=20$ for for all the image sequences of the MPI Sintel benchmark and to $\lambda_1=2$, $\lambda_2=10$, $\lambda_3=250$, $\lambda_4=4.5$ for all the image sequences of the Middlebury dataset. As a representative example (the one used to compare methods), the computation time for the \textit{Urban2} sequence of the Middlebury benchmark is 27 minutes on a Intel Xeon laptop with 2.20GHz clock speed and 64Gb RAM. Nevertheless, the first step of AggregFlow can be massively parallelized, which should lead to a far less computation cost with a GPU implementation for instance. Most of the computation time is consumed in the patch correspondence sub-step for the largest patch size ($106\times 106$ pixels). The determination of the matching variable $m$ is performed with patches of size $11\times 11$.

\subsection{Quantitative results on computer vision benchmarks}
We have evaluated AggregFlow on the two most representative benchmarks for optical flow: MPI Sintel flow dataset\footnote{http://sintel.is.tue.mpg.de/} \cite{Butler12} and Middlebury flow dataset\footnote{http://vision.middlebury.edu/flow/} \cite{Baker11}, which offer different and complementary challenges. We have retained the Endpoint Error measure (EPE) for quantitative evaluation. Results of \cite{Xu12} and \cite{Weinzaepfel13} reported in Table
\ref{table:training_results}, Fig.\ref{fig::visual_results} and Fig.\ref{fig::visual_results_mid} have been obtained with the public codes provided by the
authors\footnote{http://www.cse.cuhk.edu.hk/~leojia/projects/flow/}$^,$\footnote{http://lear.inrialpes.fr/src/deepmatching/}.\\

\noindent \textit{MPI Sintel flow dataset}~~
Sequences of the most recent MPI Sintel benchmark \cite{Butler12} are characterized by long-range motion, motion blur, non-rigid motion, and wide occluded
areas. Methods are evaluated on two versions of the sequences named {\it Clean} and {\it Final}. The Final version adds motion and defocus blur along with
atmospheric effects like fog on some sequences. We reproduce in Tables \ref{table:sintel_clean} and \ref{table:sintel_final} public results of the top ranked
published methods, which are presently (at paper submission date, June 4, 2014) available on the MPI Sintel website. Results are analyzed through several
indicators:
``EPE all'' is the average EPE on all the sequences; ``EPE matched'' and ``EPE unmatched'' restrict the error measure respectively to regions that remain visible in adjacent frames (non-occluded pixels) and to regions that are visible only in one of two adjacent frames (occluded pixels); ``d0-10'' denotes EPE over regions closer than 10 pixels to the nearest occlusion boundary, and thus reveals the ability to recover motion discontinuities; ``s40+'' denotes EPE over regions with velocities larger than 40 pixels per frame. Methods are ranked regarding their EPE all. Visual comparison with results supplied by \cite{Weinzaepfel13} and \cite{Xu12} on training sequences (i.e., MPI Sintel sequences provided with ground truth) is available in Fig.\ref{fig::visual_results}.

As for the \textit{Clean} subset, our method AggregFlow ranks first over the published methods. The most significant improvement is obtained on the  unmatched category, which emphasizes the efficiency of our occlusion framework. AggregFlow is ranked second for the d0-10 metric, which demonstrates its capacity to recover motion discontinuities as confirmed by results displayed in Fig.\ref{fig::visual_results}. First, it is due to the robust affine estimation of  the motion candidates able to capture locally dominant motion in case of two or even several motions present inside patches, which preserves motion discontinuities. It is also made successful by the efficient occlusion module, which allows us to moderate the need for motion field regularization.
Indeed, missing information in occluded regions is usually tackled by imposing high regularization with the result of oversmoothing the rest of the motion field (see motion fields computed with DeepFlow \cite{Weinzaepfel13} in Fig.\ref{fig::visual_results}). In case of very large displacements (s40+ metric), all the first five methods (AggregFlow, \cite{Weinzaepfel13,Xu12,Bao14,Leordeanu13}) somehow integrate feature matching in their motion estimation process
to capture the largest deformations. The top rank of AggregFlow demonstrates the efficiency of the aggregation framework for integrating feature
matching.

As for the \textit{Final} version AggregFlow is ranked second in terms of EPE all. The slight decreasing in performance compared to the Clean subset is mainly due to errors caused by the added fog effect in the two \textit{ambush} sequences. As emphasized in \cite{Bao14}, local intensity-based displacement computation tends to capture the motion of the fog rather than the movement of objects appearing in transparency. As our candidates estimation is local, it is subject to this limitation. Global variational approaches are able to diffuse motion estimates in these regions and are consequently better suited for this kind of situations. Despite this shortcoming, our method still yields significant improvement in unmatched regions and on motion discontinuities. One solution to improve results in fog regions would be to incorporate a more sophisticated feature correspondence technique as the ones proposed in \cite{Leordeanu13,Weinzaepfel13}.\\

\begin{table}[!t]
 \centering
 \caption{Results on the MPI Sintel Clean test subset.}
~\\
 \begin{tabular}{l|c|c|c|c|c}
 & EPE & EPE & EPE & d0-10 & s40+ \\
 & all & matched & unmatched &  & 	 \\
\hlinewd{1pt}
 { {\bf AggregFlow}} 						& {\bf 4.754} & {\bf 1.694} & {\bf 29.685} & 3.705 & {\bf 31.184}  \\
 { {DeepFlow \cite{Weinzaepfel13}}}			& 5.377 & 1.771 & 34.751 & 4.519 & 33.701   \\
 { {MDP-Flow2 \cite{Xu12}}} 				& 5.837 & 1.869 & 38.158 & {\bf 3.210} & 39.459  \\
 { {EPPM \cite{Bao14}}} 					& 6.494 & 2.675 & 37.632 & 4.997 & 39.152  \\
 { {S2D-Matching \cite{Leordeanu13}}} 		& 6.510 & 2.792 & 36.785 & 5.523 & 44.187  \\
 { {Classic+NLP \cite{Sun14}}} 				& 6.731 & 2.949 & 37.545 & 5.573 & 45.290  \\
 { {FC-2Layers-FF \cite{Sun12}}}			& 6.781 & 3.053 & 37.144 & 5.841 & 45.962  \\
 { {MLDP-OF \cite{Mohamed14}}}				& 7.297 & 3.260 & 40.183 & 5.581 & 51.146  \\
 \end{tabular}
  \label{table:sintel_clean}
 \end{table}

\begin{table}[!t]
 \centering
 \caption{Results on the MPI Sintel Final test subset.}
~\\
 \begin{tabular}{l|c|c|c|c|c}
 & EPE & EPE & EPE & d0-10 & s40+ \\
 & all & matched & unmatched &  & 	 \\
\hlinewd{1pt}
 {{DeepFlow \cite{Weinzaepfel13}}}	& {\bf 7.212} & {\bf 3.336} & 38.781 & 5.650 & {\bf 44.118}   \\
 {{\bf AggregFlow}} 						& 7.329 & 3.696 & {\bf 36.929} & {\bf 5.538} & 44.858  \\
 {{S2D-Matching	\cite{Leordeanu13}}} 	& 7.872 & 3.918 & 40.093 & 5.975 & 48.782  \\
 {{FC-2Layers \cite{Sun12}}} 		& 8.137 & 4.261 & 39.723 & 6.537 & 51.349  \\
 {{MLDP-OF \cite{Mohamed14}}} 				& 8.287 & 4.165 & 41.905 & 6.345 & 50.540  \\
 {{Classic+NLP \cite{Sun14}}}			& 8.291 & 4.287 & 40.925 & 6.520 & 51.162  \\
 {{EPPM \cite{Bao14}}} 				& 8.377 & 4.286 & 41.695 & 6.556 & 49.083  \\
 {{MDP-Flow2 \cite{Xu12}}} 			& 8.445 & 4.150 & 43.430 & 5.703 & 50.507  \\
 \end{tabular}
  \label{table:sintel_final}
 \end{table}

\begin{table}[!t]
 \centering
 \caption{Results on the Middlebury benchmark for the same set of methods.}
~\\
 \begin{tabular}{l|c|c}
 & EPE all & Avg. rank \\
\hlinewd{1pt}
 {{MDP-Flow2 \cite{Xu12}}} 			& {\bf 0.245}	& {\bf 7.8}   \\
 {{FC-2Layers-FF \cite{Sun12}}} 		& 0.283 & 19.3  \\
 {{Classic+NL \cite{Sun14}}}			& 0.319 & 27.1   \\
 {{EPPM \cite{Bao14}}} 				& 0.329 & 32.6   \\
 {{\bf AggregFlow}} 					& 0.339 & 35.9   \\
 {{MLDP-OF \cite{Mohamed14}}} 			& 0.349 & 32.6   \\
 {{S2D-Matching \cite{Leordeanu13}}} 	& 0.347 & 34.6   \\
 {{DeepFlow \cite{Weinzaepfel13}}}		& 0.416 & 48.8   \\
 \end{tabular}
  \label{table:middlebury}
 \end{table}

\begin{table*}[t]
\centering
\caption{Results on the MPI Sintel training subset. Scores correspond to the EPE all metric}
~\\
\begin{tabular}{l|c|c|c|c}
& {{\bf AggregFlow}} & {{\bf AggregFlow}} & {{DeepFlow \cite{Weinzaepfel13}}} & {{MDP-Flow2 \cite{Xu12}}}\\
& ~ & {{\bf w/o occlusion}} & & ~\\
\hlinewd{1pt}
{\it ambush\_2} & \bf ~5.632  &  ~9.456 & 14.743 & 12.083\\
{\it ambush\_4} & \bf 11.923 & 16.515 & 14.647 & 15.570\\
{\it ambush\_5} & \bf ~5.042  &  ~5.500 &  ~8.333 &  ~6.591\\
{\it ambush\_6} & \bf ~5.854  &  ~6.251 &  ~9.928 &  ~8.466\\
{\it market\_5} & \bf ~9.957  & 11.958 & 15.056 & 12.816\\
{\it market\_6} & \bf ~3.626  &  ~4.547 &  ~6.606 &  ~5.384\\
{\it cave\_2}   & \bf ~6.029  &  ~8.228 & 10.082 &  ~8.347\\
{\it cave\_4}   & \bf ~3.706  &  ~4.185 &  ~4.234 &  ~3.815\\
{\it temple\_3} & \bf ~5.875  &  ~8.314 & 11.895 &  ~9.011\\
\hline
{\bf Average}   & \bf ~6.002 & ~8.417 & 10.614 & ~9.120\\ 
\end{tabular}
\label{table:training_results}
\end{table*}

\noindent \textit{Middlebury dataset}~~
The Middlebury benchmark is composed of sequences with small displacements, where the main challenge is to be able to recover both complex smooth deformation,
motion discontinuities and motion details. Table \ref{table:middlebury} reproduces public results presently available (at paper submission date, June 4, 2014)
for the same methods as those taken for comparison on the MPI Sintel benchmark. Visual comparative results are displayed in Fig.\ref{fig::visual_results_mid}. It can be observed that the EPE values, together with the differences between methods, are much lower than for the MPI Sintel dataset. The average EPE score computed over the considered methods is equal to 6.22 for the MPI Sintel Clean subset and to 0.327 for the Middlebury dataset, with respective variance of 0.613 and 0.0025. We also provide the average rank over the 8 test sequences for each method which is the metric used for global ranking on the Middlebury website.

On the whole Middlebury benchmark, AggregFlow is ranked 38 over 97 tested methods in terms of average rank on the results (evaluated with the average endpoint error on the sequence) obtained for the eight test sequences. Notwithstanding, it is still very close to the ranked two MDP-Flow2 method \cite{Xu12} in terms of EPE metric, knowing that the top ranked published method OFLAF \cite{Kim13} has an average rank of 6.8 and an EPE all of 0.197 (OFLAF method was not tested on the MPI Sintel benchmark). Visual results reported in Fig.\ref{fig::visual_results_mid} confirm the tightness of performance. In particular, the preservation of motion discontinuities with AggregFlow is more satisfying than with the DeepFlow method \cite{Weinzaepfel13}. These results also show that AggregFlow is competitive for recovering motion details in addition to the large velocities of the MPI Sintel benchmark.

\subsection{Occlusion handling}
\begin{figure}[htbp]
  \centering
   $\begin{array}{c@{\hspace{2pt}}c@{\hspace{2pt}}c@{\hspace{2pt}}c@{\hspace{2pt}}c}
\includegraphics[width=115pt]{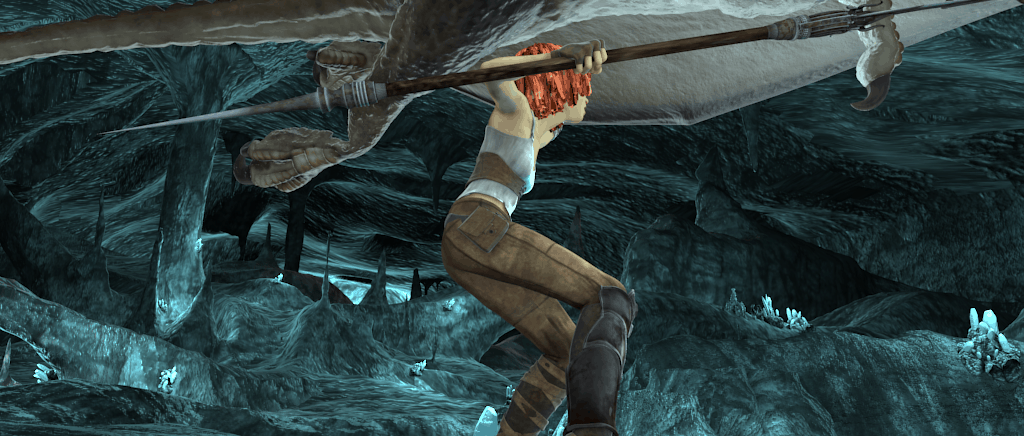} &
\includegraphics[width=115pt]{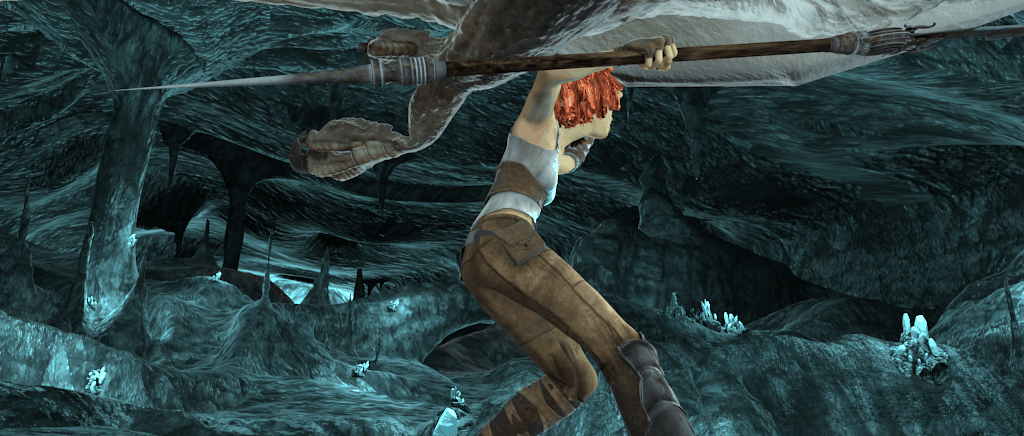} &
 \includegraphics[width=115pt]{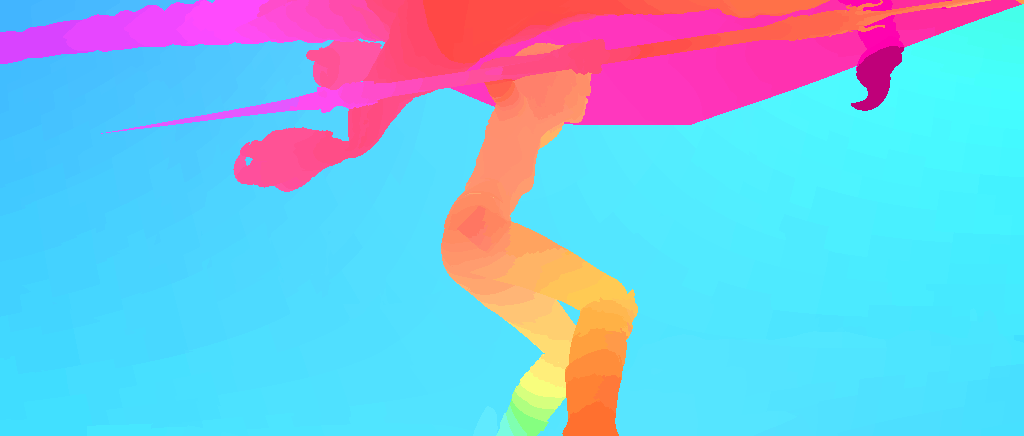} &
 \includegraphics[width=115pt]{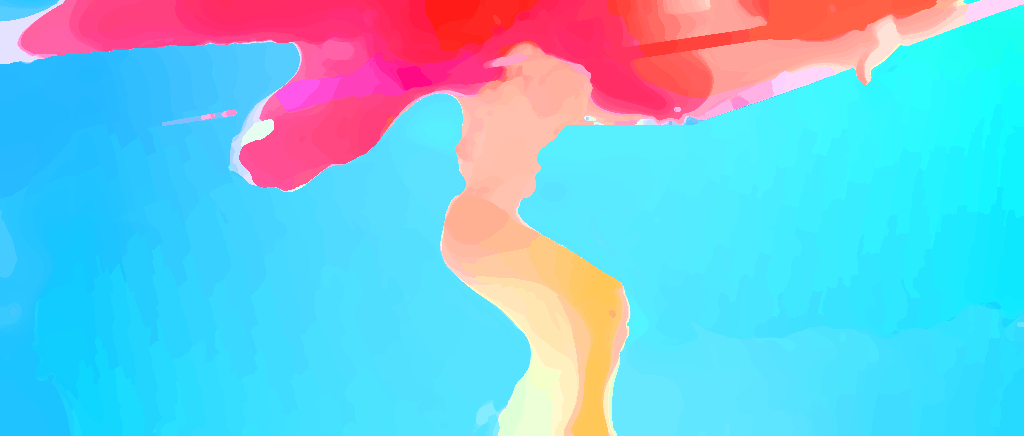} \\
  \mbox{\small \small {\it cave\_2} - $I_1$} & \mbox{\small \small {\it cave\_2} - $I_2$} & \mbox{\small \small Ground truth $\w$}  & \mbox{\small \small AggregFlow} \\
 \includegraphics[width=115pt]{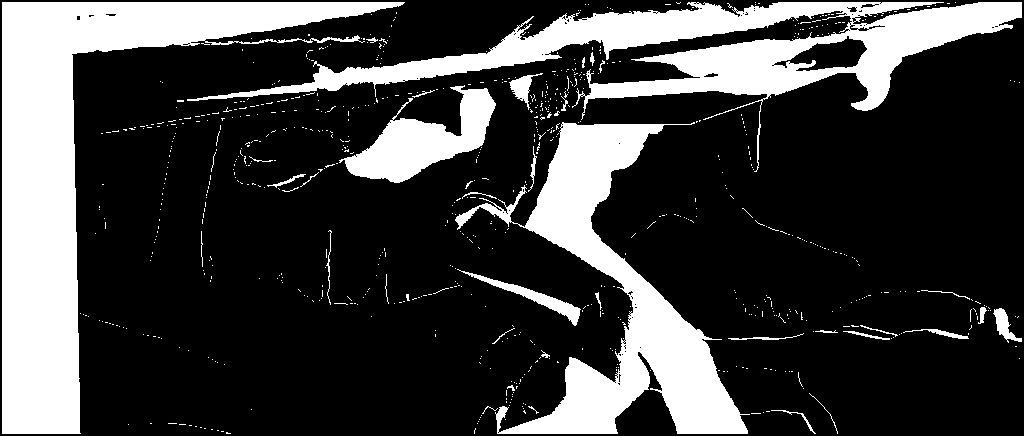} &
  \includegraphics[width=115pt]{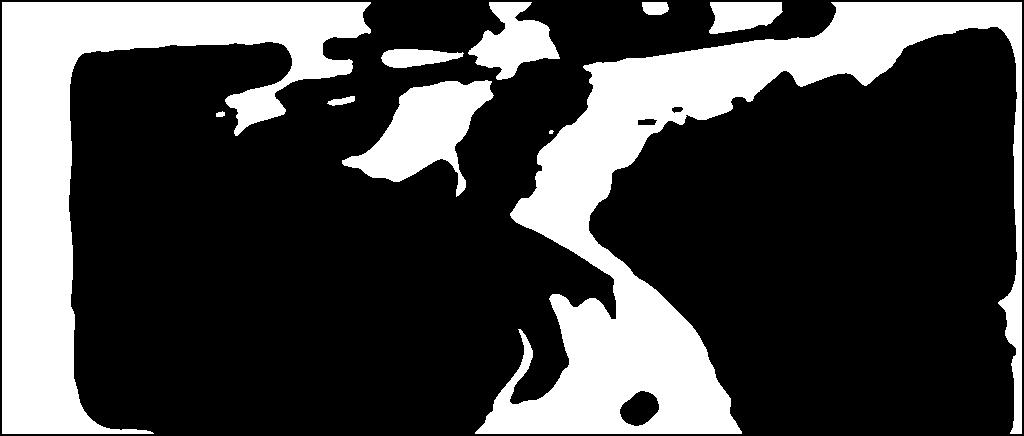} &
 \includegraphics[width=115pt]{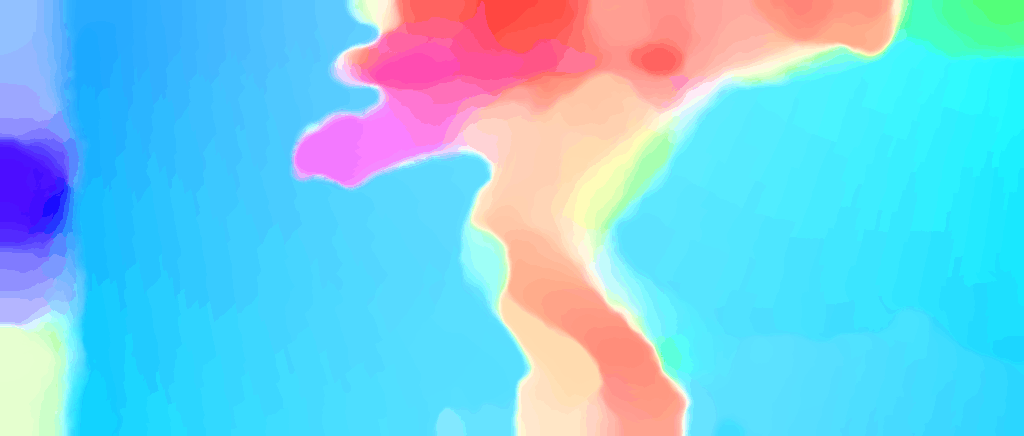} &
 \includegraphics[width=115pt]{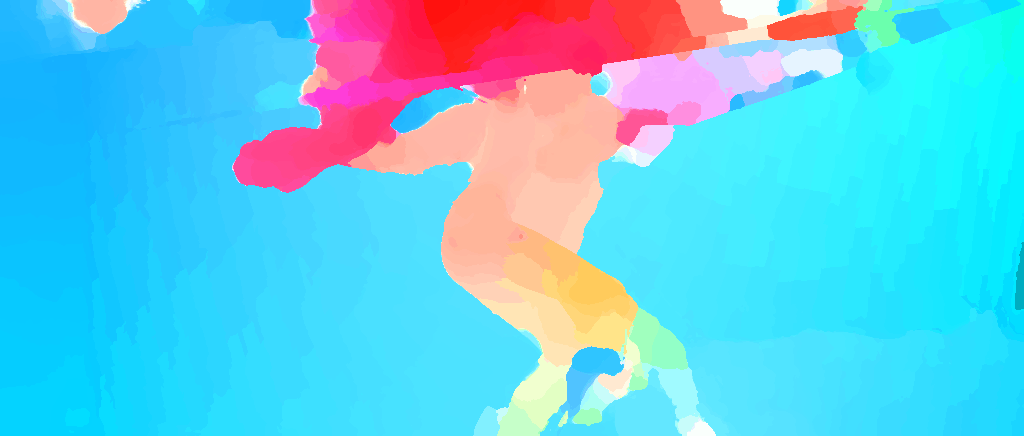} \\
 \mbox{\small \small Ground truth $o$} & \mbox{\small \small AggregFlow} & \mbox{\small \small DeepFlow \cite{Weinzaepfel13}} & \mbox{\small \small MDP-Flow2 \cite{Xu12}} \\[7pt]

 \includegraphics[width=115pt]{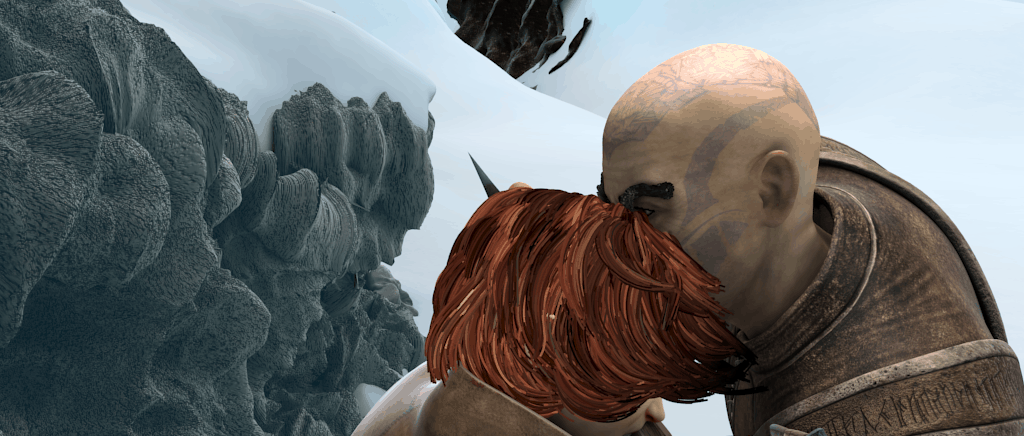} &
 \includegraphics[width=115pt]{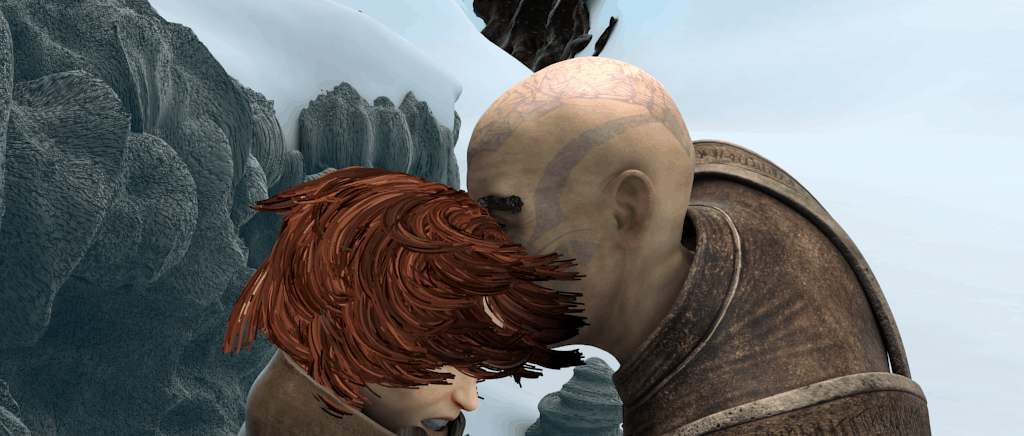} &
 \includegraphics[width=115pt]{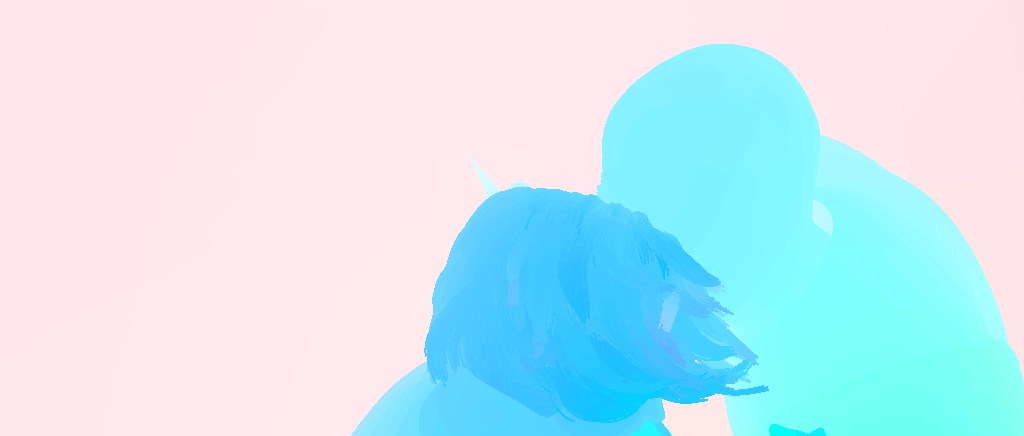} &
 \includegraphics[width=115pt]{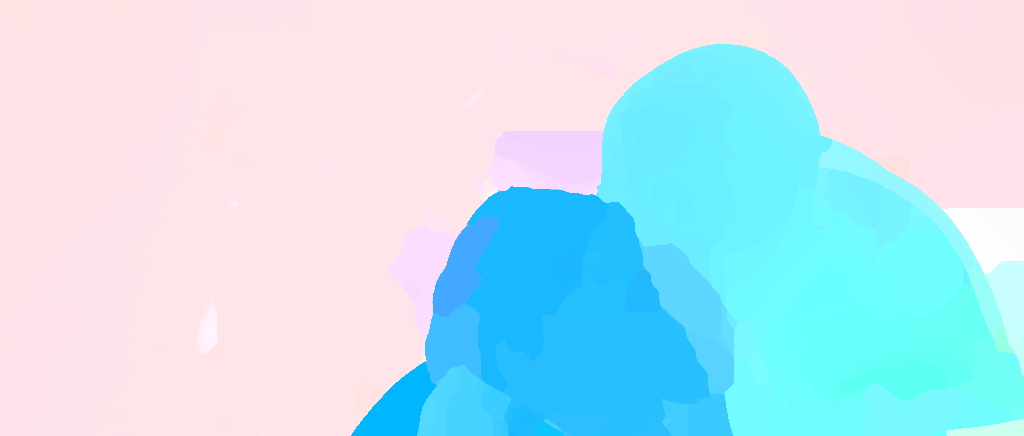}\\
  \mbox{\small \small {\it ambush\_5} - $I_1$} & \mbox{\small \small {\it ambush\_5} - $I_2$} & \mbox{\small \small Ground truth $\bf w$}  & \mbox{\small \small AggregFlow} \\
 \includegraphics[width=115pt]{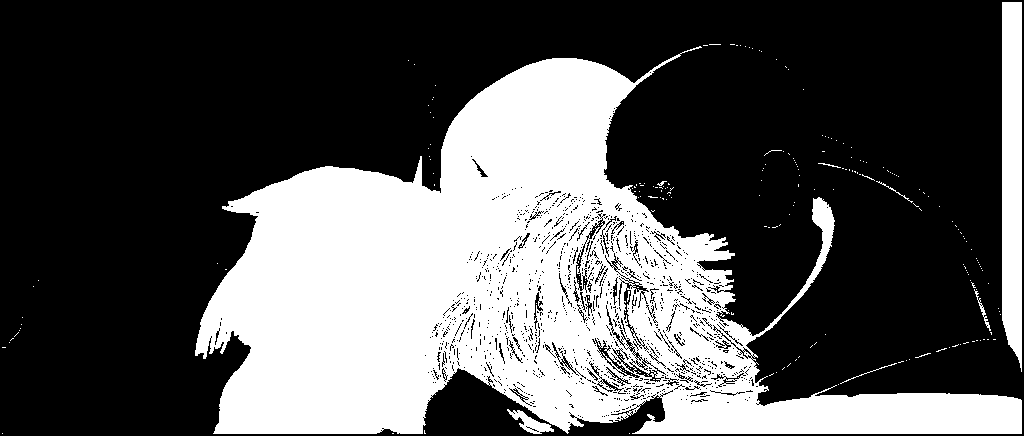} &
  \includegraphics[width=115pt]{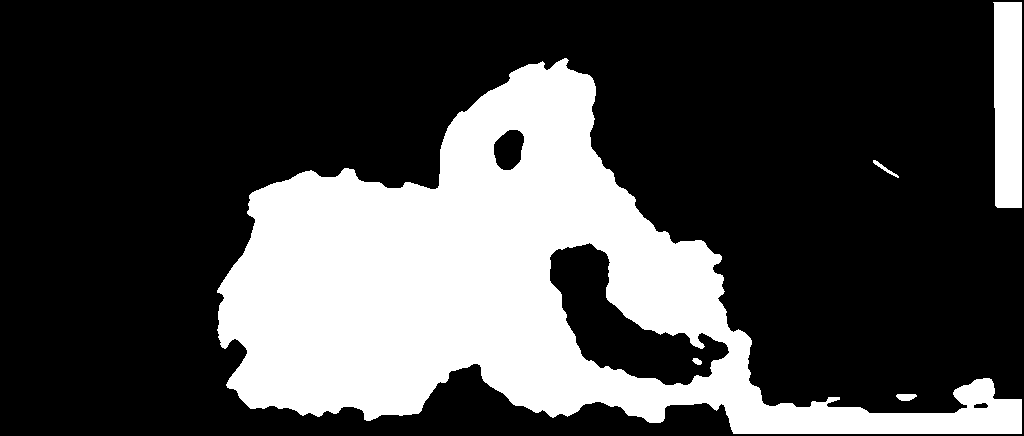} &
 \includegraphics[width=115pt]{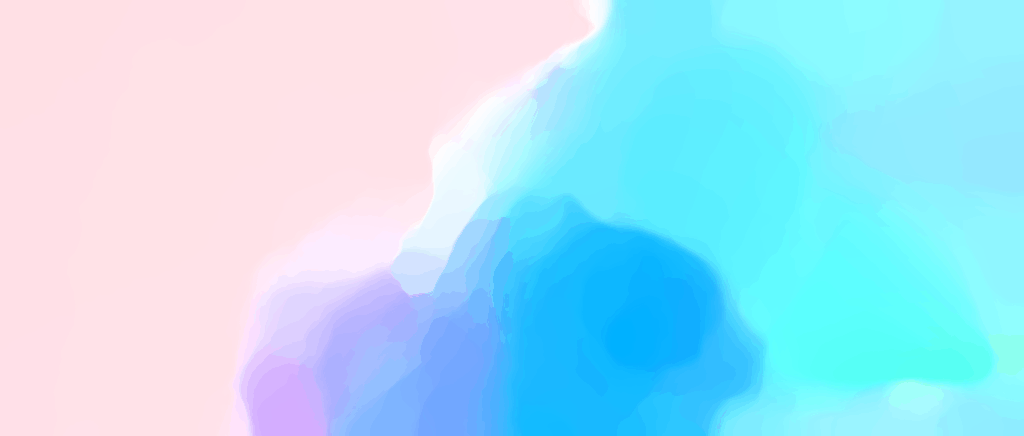} &
 \includegraphics[width=115pt]{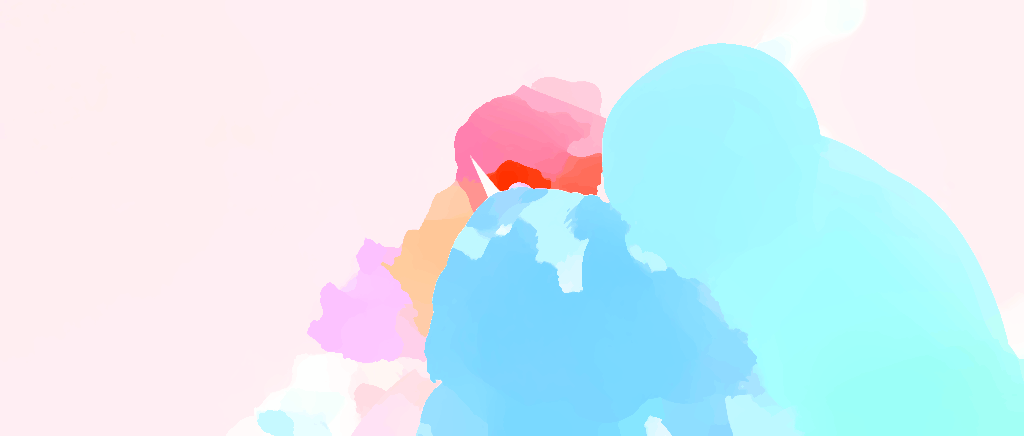} \\
 \mbox{\small \small Ground truth $o$} & \mbox{\small \small AggregFlow} & \mbox{\small \small DeepFlow \cite{Weinzaepfel13}} & \mbox{\small \small MDP-Flow2 \cite{Xu12}} \\[7pt]

 \includegraphics[width=115pt]{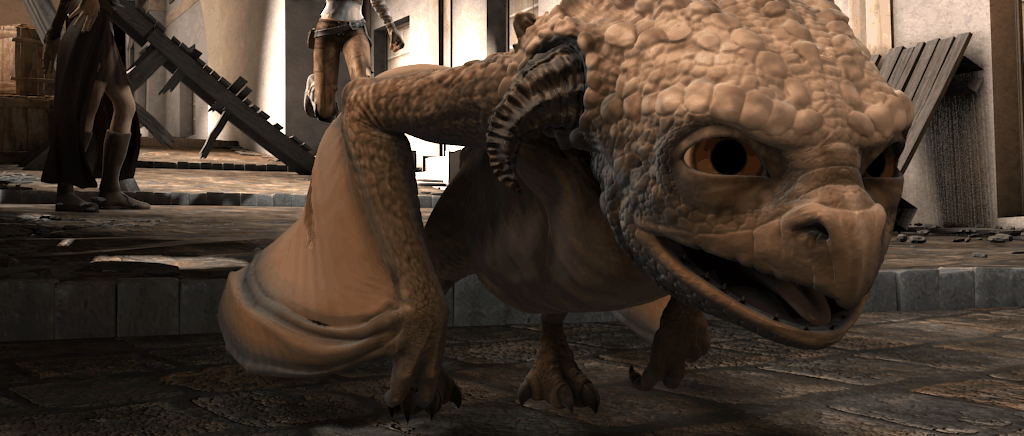} &
 \includegraphics[width=115pt]{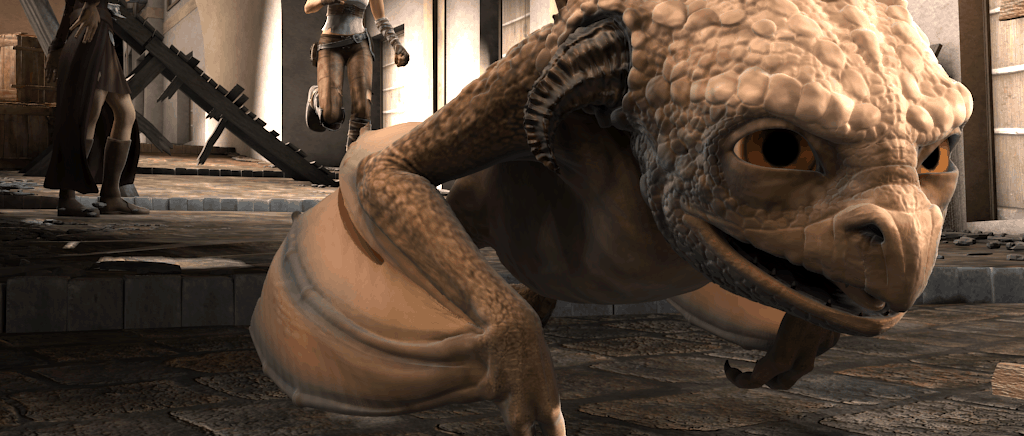} &
 \includegraphics[width=115pt]{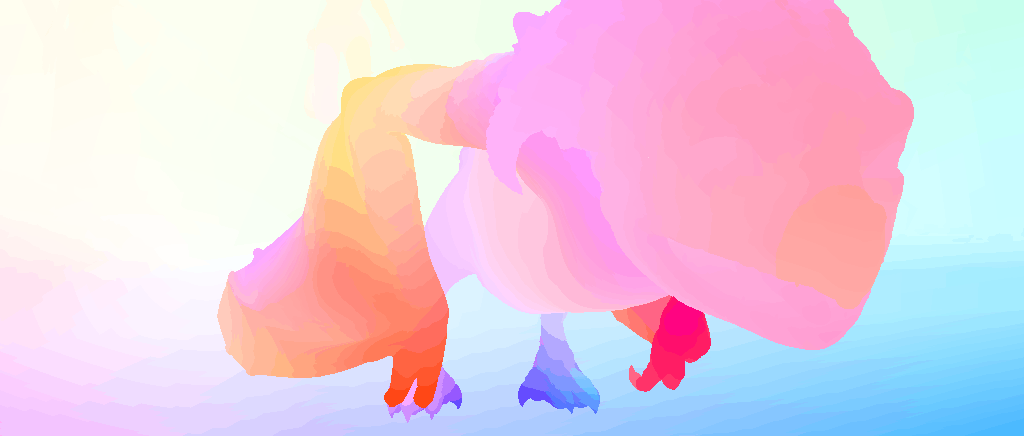} &
 \includegraphics[width=115pt]{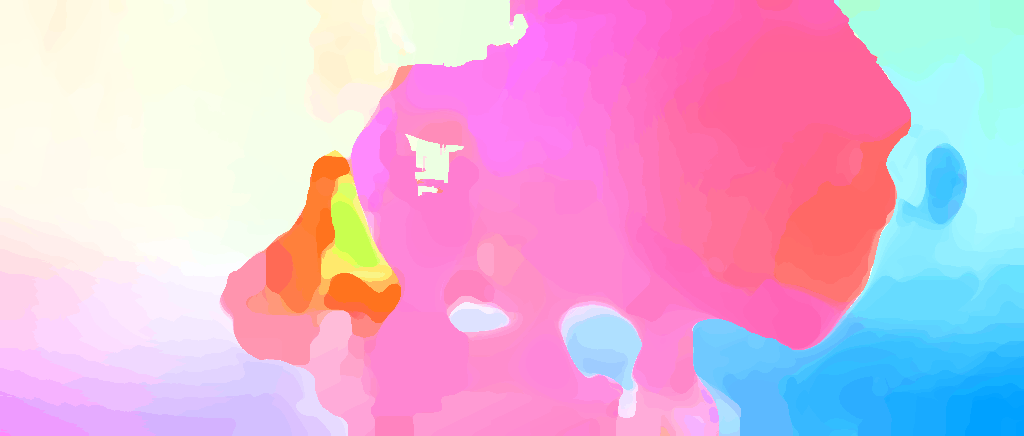} \\
  \mbox{\small \small {\it market\_5} - $I_1$} & \mbox{\small \small {\it market\_5} - $I_2$} & \mbox{\small \small Ground truth $\w$}  & \mbox{\small \small AggregFlow} \\
 \includegraphics[width=115pt]{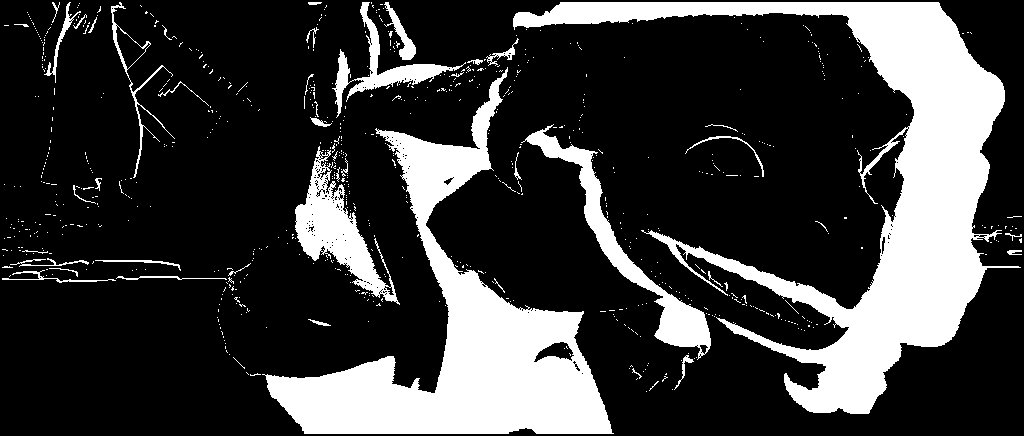} &
  \includegraphics[width=115pt]{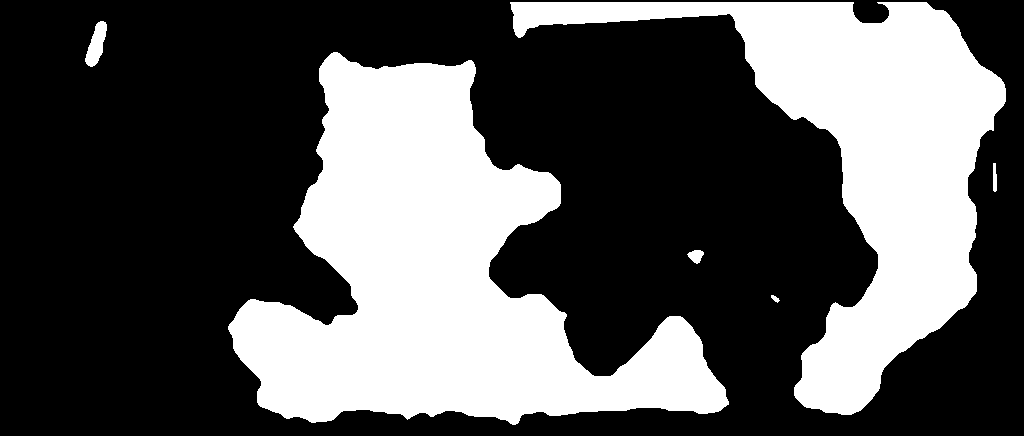} &
 \includegraphics[width=115pt]{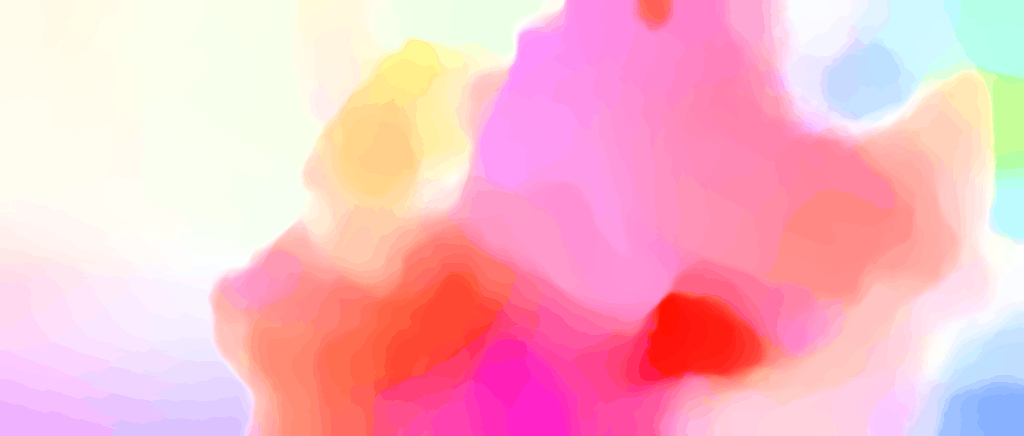} &
 \includegraphics[width=115pt]{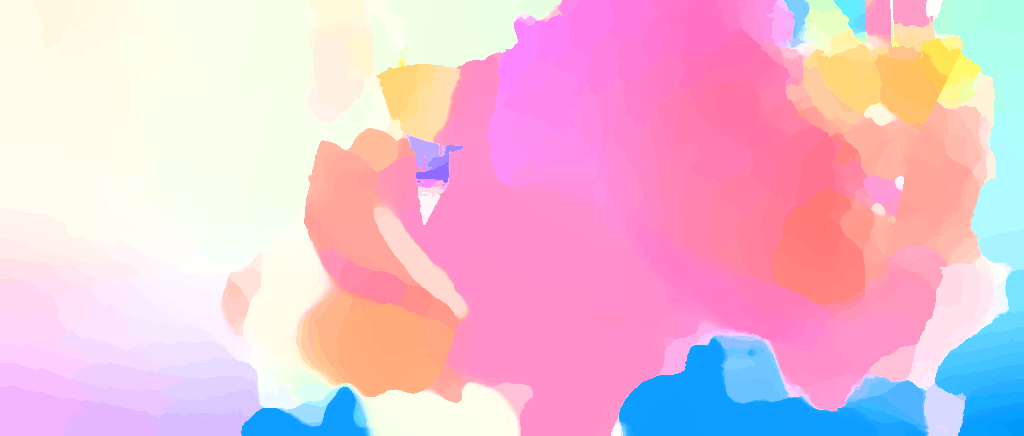} \\
 \mbox{\small \small Ground truth $o$} & \mbox{\small \small AggregFlow} & \mbox{\small \small DeepFlow \cite{Weinzaepfel13}} & \mbox{\small \small MDP-Flow2 \cite{Xu12}} \\[7pt]

 \includegraphics[width=115pt]{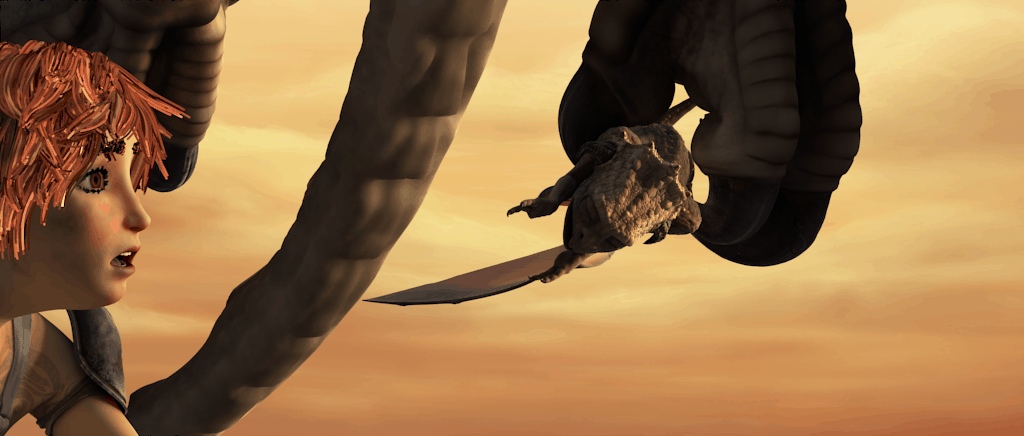} &
 \includegraphics[width=115pt]{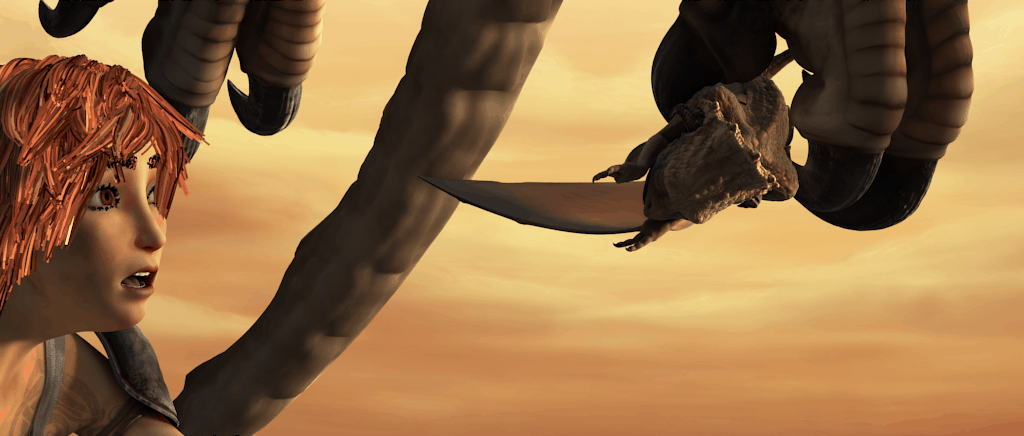} &
 \includegraphics[width=115pt]{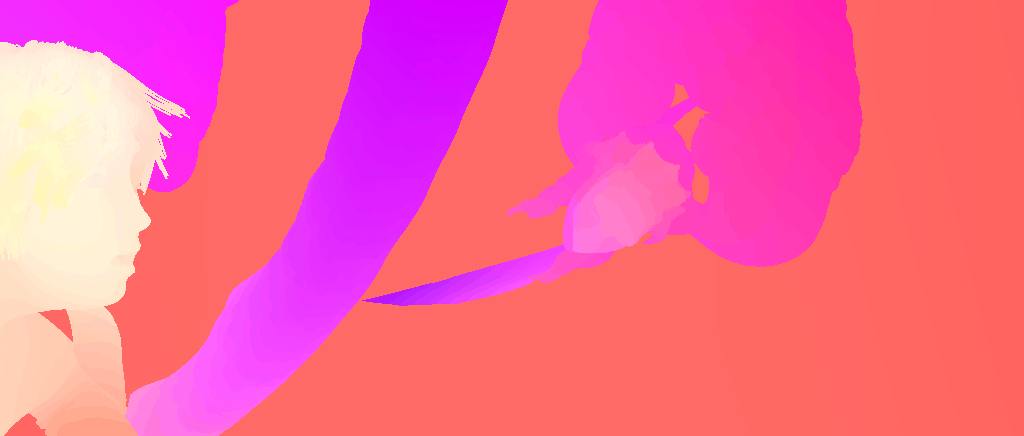} &
 \includegraphics[width=115pt]{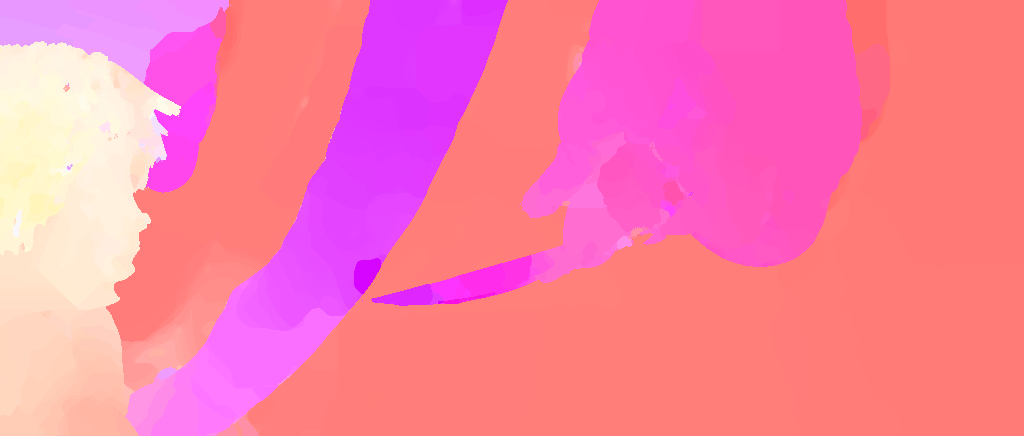} \\
  \mbox{\small \small {\it temple\_3} - $I_1$} & \mbox{\small \small {\it temple\_3} - $I_2$} & \mbox{\small \small Ground truth $\w$}  & \mbox{\small \small AggregFlow} \\
 \includegraphics[width=115pt]{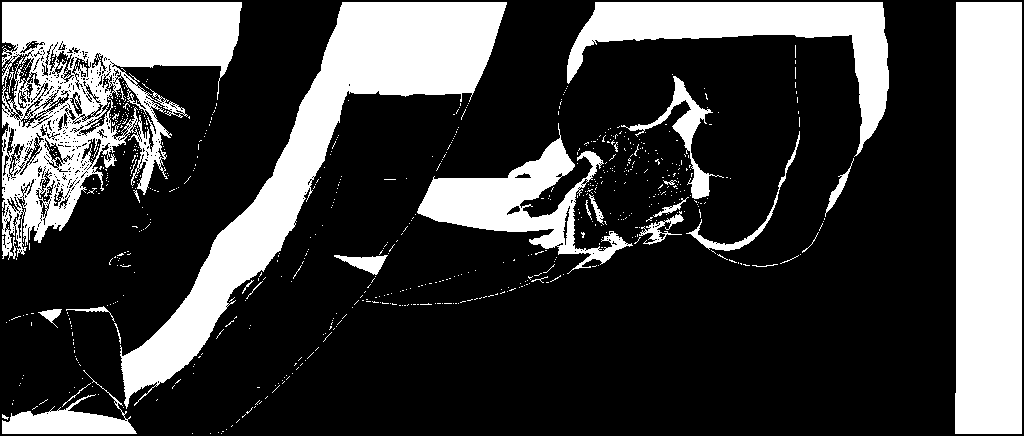} &
  \includegraphics[width=115pt]{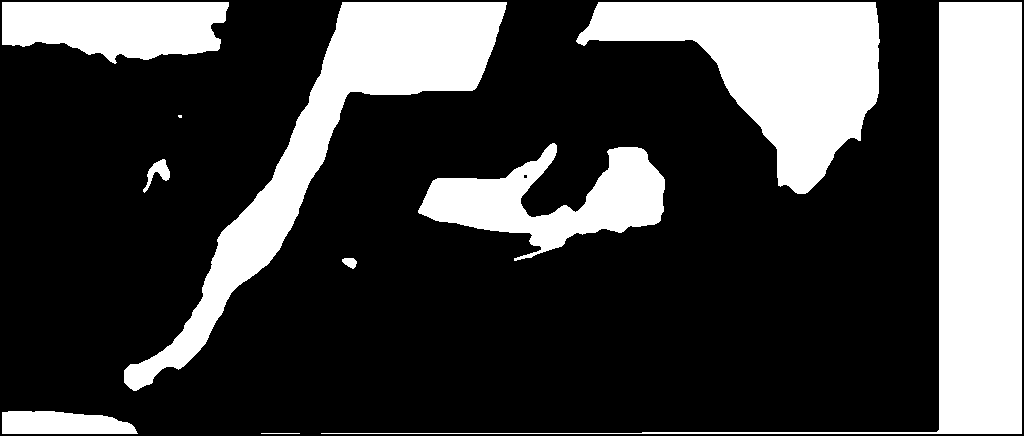} &
 \includegraphics[width=115pt]{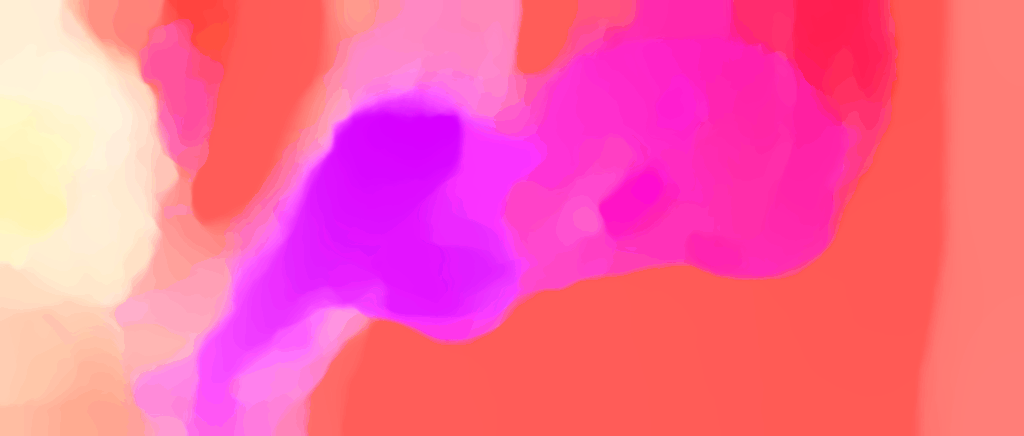} &
 \includegraphics[width=115pt]{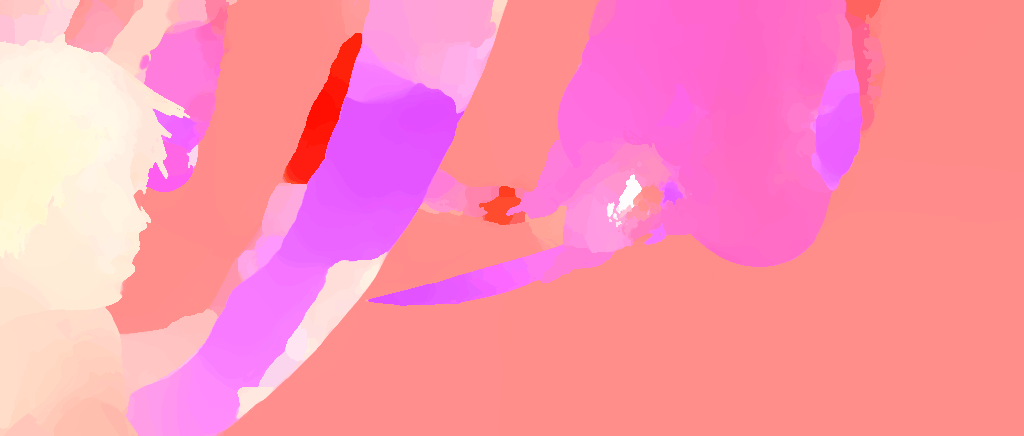} \\
 \mbox{\small \small Ground truth $o$} & \mbox{\small \small AggregFlow} & \mbox{\small \small DeepFlow \cite{Weinzaepfel13}} & \mbox{\small \small MDP-Flow2 \cite{Xu12}} \\

   \end{array}$
  \caption{Comparative evaluation with \cite{Weinzaepfel13} and \cite{Xu12} on several sequences of the MPI Sintel dataset. Every first row from left to right: successive input images, ground truth motion field, motion field computed with AggregFlow. Every second row from left to right: ground truth occlusion, occlusion map computed with AggregFlow, motion fields computed with DeepFlow \cite{Weinzaepfel13} and MDP-Flow2 \cite{Xu12}.}
\label{fig::visual_results}

  \end{figure}

As aforementioned the impact of our occlusion framework on optical flow estimation is demonstrated by the EPE unmatched metric scores obtained on the MPI
Sintel benchmark (Tables \ref{table:sintel_clean} and \ref{table:sintel_final}). Results of Fig.\ref{fig::visual_results} reveal the superiority of AggregFlow in coping with occluded regions. Since the occlusion framework is composed of several elements, we detail the influence of each one in the following. 
The efficiency of the motion candidates extension in occluded regions has already been highlighted in Section \ref{sec:BCF} and Table \ref{table:bcf}
through the analysis of the Best Candidate Flow.

To evaluate the occlusion model of the aggregation step, we report in Table \ref{table:training_results} results obtained on a selection of training
sequences of the MPI Sintel benchmark with the largest displacements. We distinguish between the full AggregFlow method, and AggregFlow without the
occlusion-related terms in (\ref{eq:Eaggreg}), that is, by setting $\lambda_1=0$, $\lambda_2=0$ and $\lambda_3=0$. The improvement due to the occlusion terms is clearly significant since the average EPE is 8.417 for AggregFlow without occlusions and 6.002 for full Aggregflow. It can also be noticed that even without handling occlusion AggregFlow still performs better than competing methods. The role of the occlusion confidence map involved in the sparsity constraint (\ref{eq:occ_constraint}) has already been shown in Section \ref{sec:occ_constraint} and Fig.\ref{fig:occ_conf_final}.

\begin{figure}[t]
  \centering
   $\begin{array}{c@{\hspace{2pt}}c@{\hspace{2pt}}c@{\hspace{2pt}}c@{\hspace{2pt}}c@{\hspace{2pt}}c@{\hspace{2pt}}c}

  \includegraphics[width=76pt]{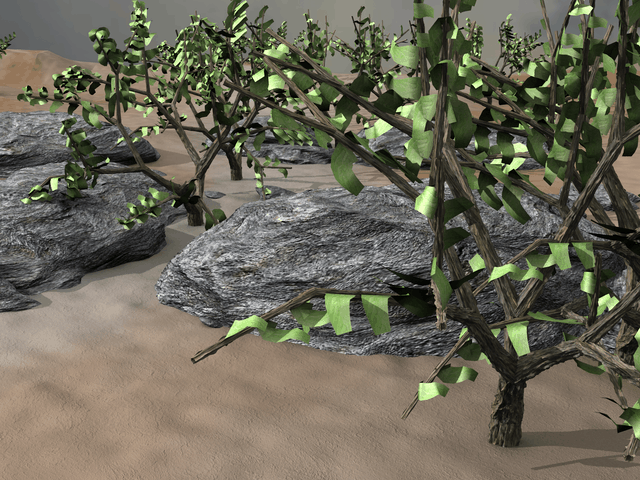} &
 \includegraphics[width=76pt]{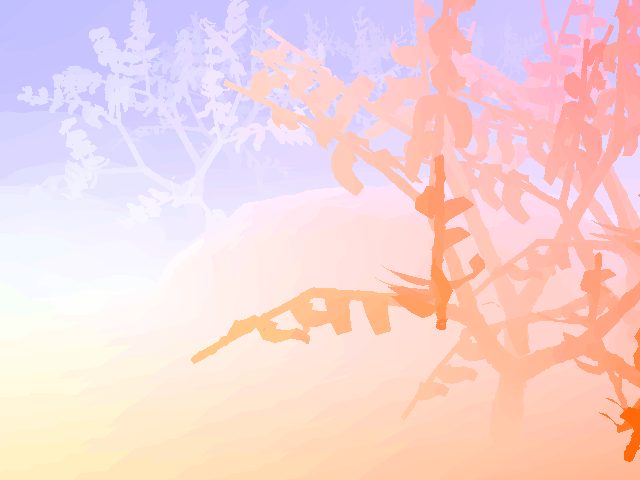} &
 \includegraphics[width=76pt]{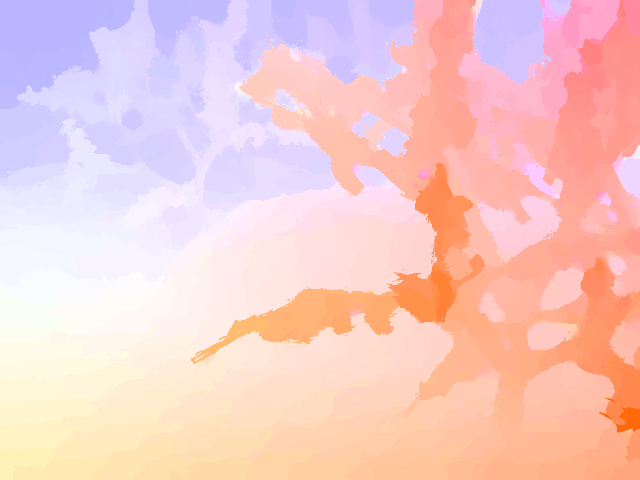} &
 \includegraphics[width=76pt]{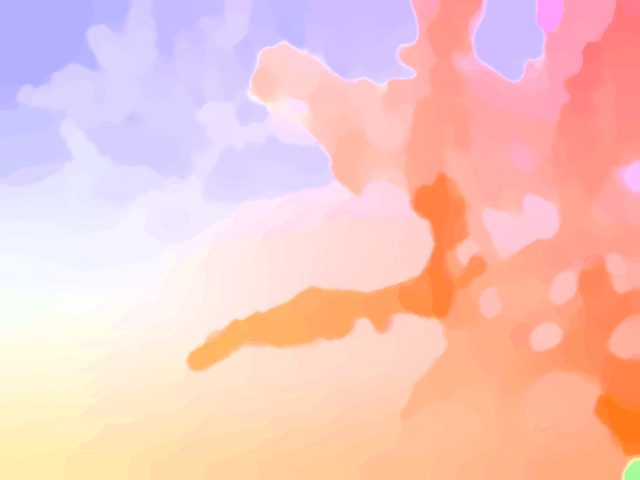} &
 \includegraphics[width=76pt]{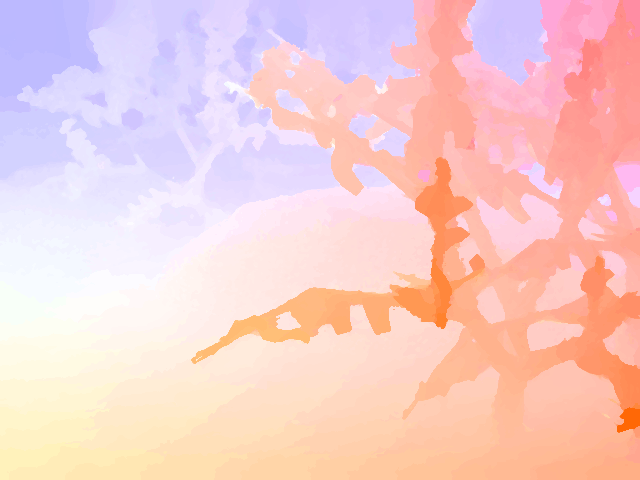} &
  \includegraphics[width=76pt]{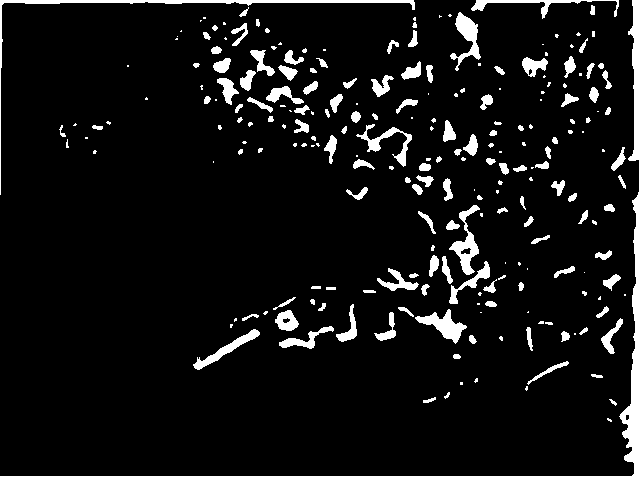} \\

 \includegraphics[width=76pt]{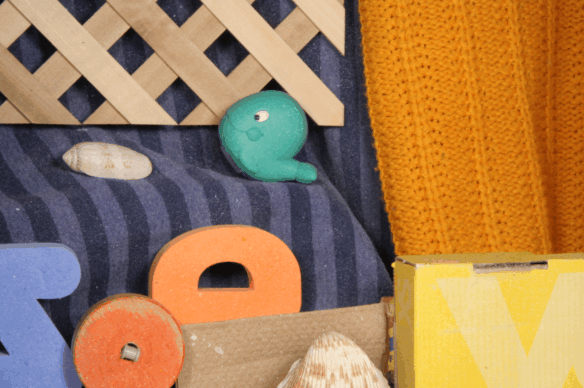} &
 \includegraphics[width=76pt]{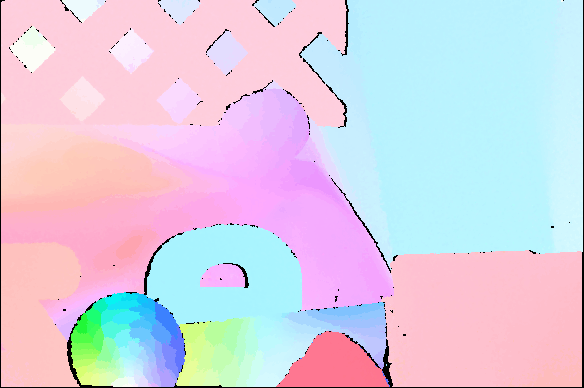} &
 \includegraphics[width=76pt]{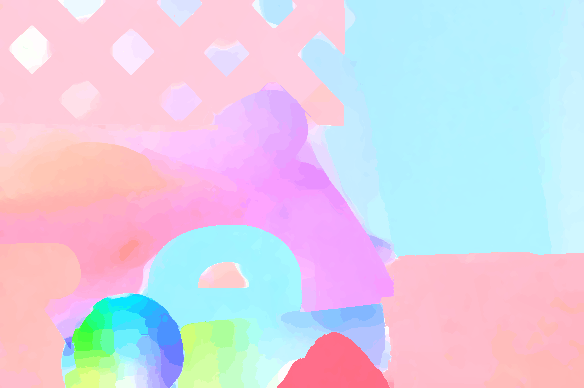} &
 \includegraphics[width=76pt]{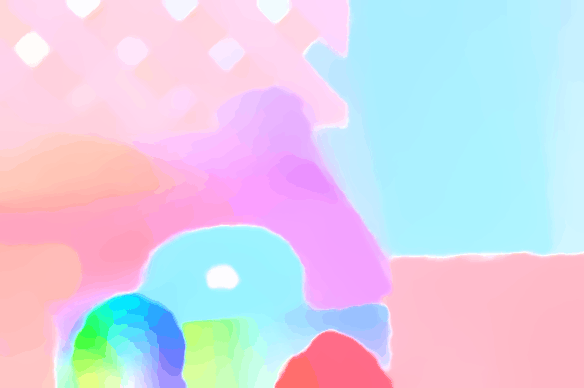} &
 \includegraphics[width=76pt]{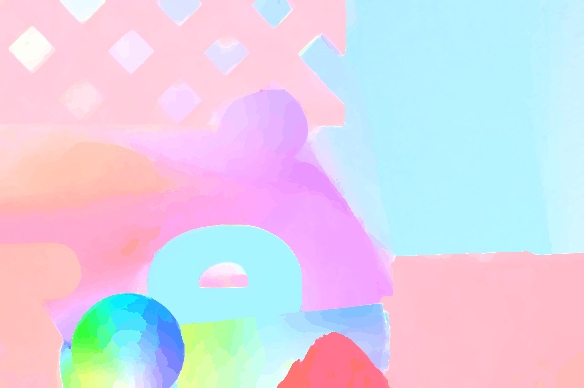} &
  \includegraphics[width=76pt]{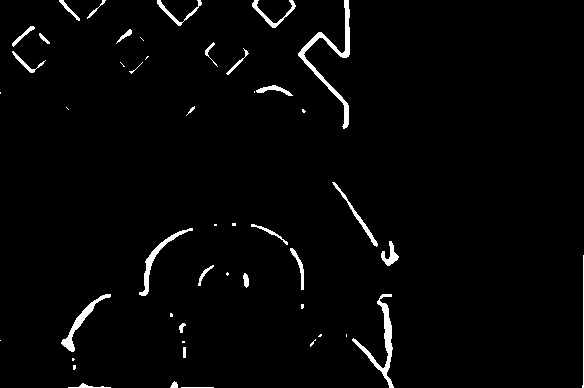} \\

 \includegraphics[width=76pt]{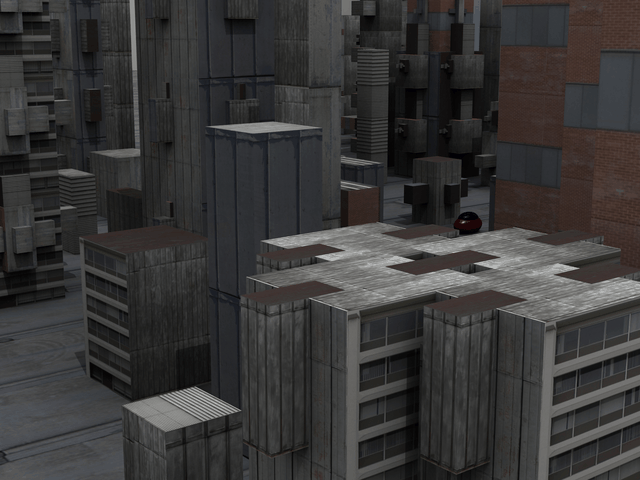} &
 \includegraphics[width=76pt]{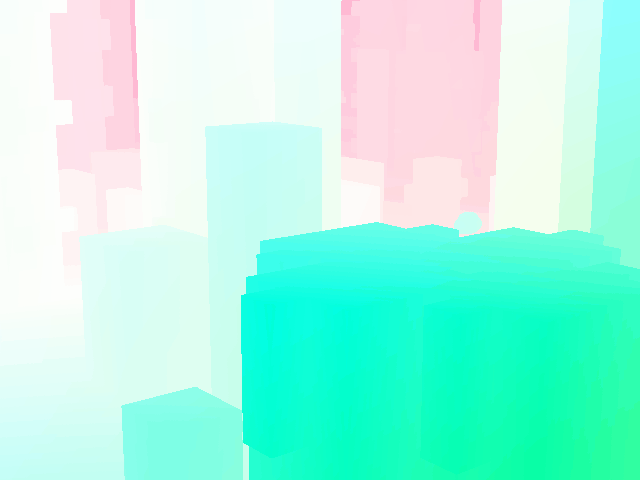} &
 \includegraphics[width=76pt]{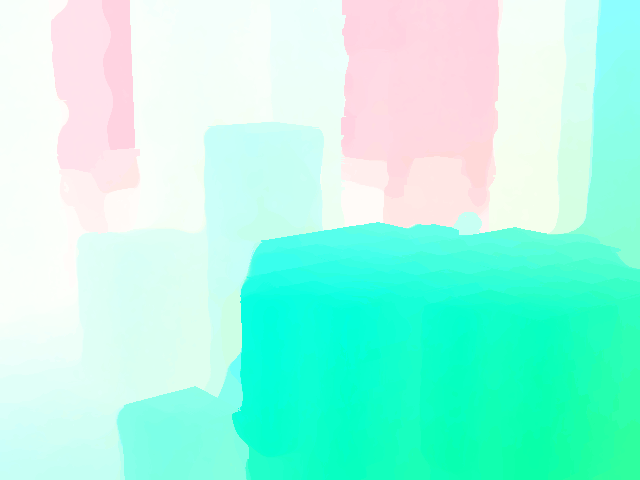} &
 \includegraphics[width=76pt]{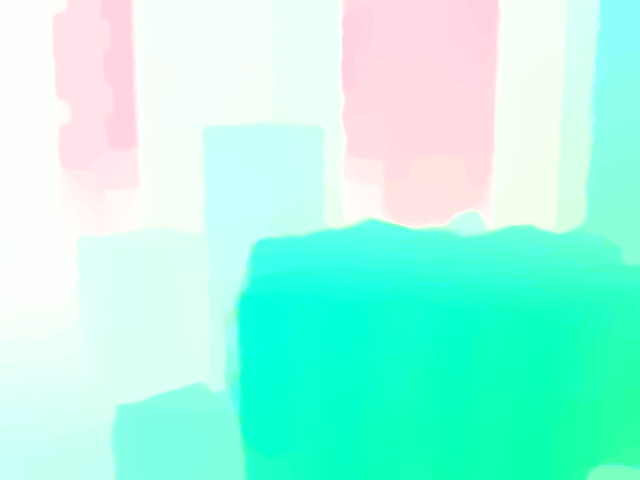} &
 \includegraphics[width=76pt]{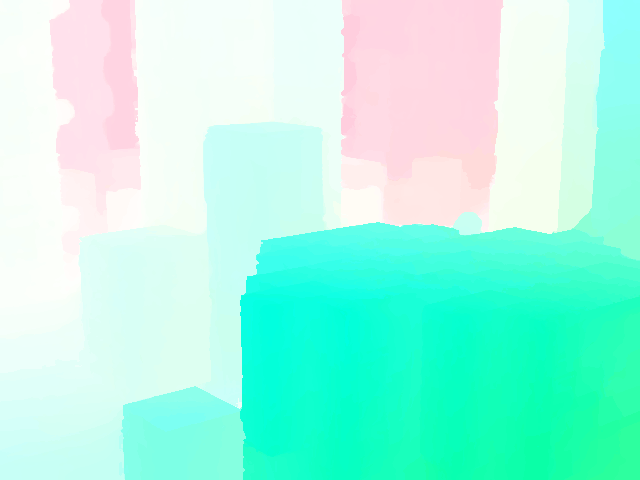} &
 \includegraphics[width=76pt]{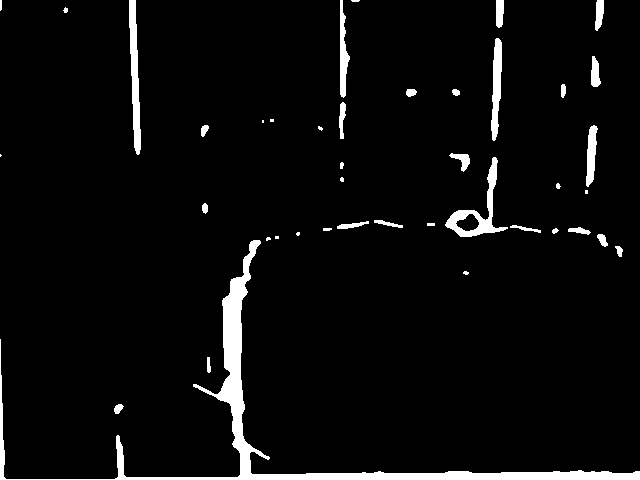} \\

 \includegraphics[width=76pt]{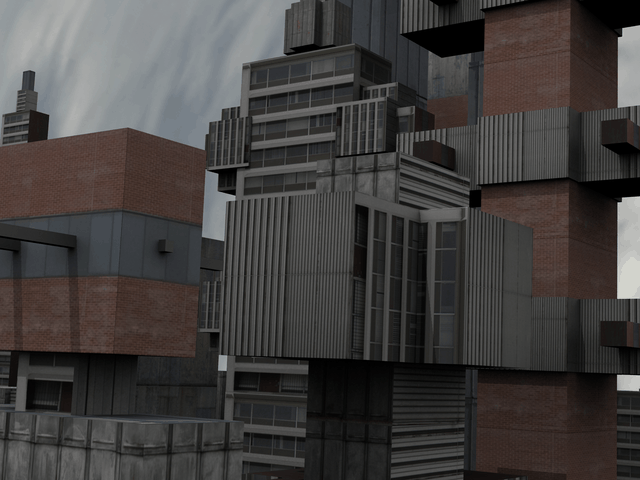} &
 \includegraphics[width=76pt]{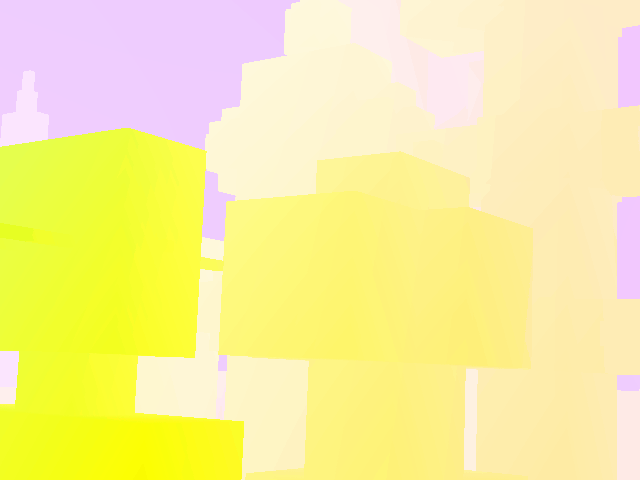} &
 \includegraphics[width=76pt]{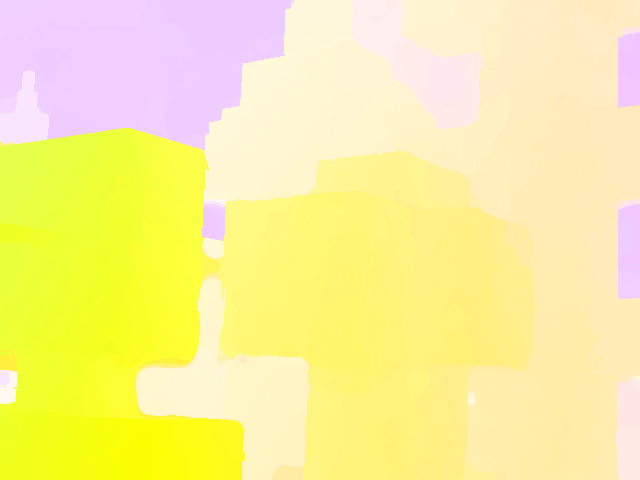} &
 \includegraphics[width=76pt]{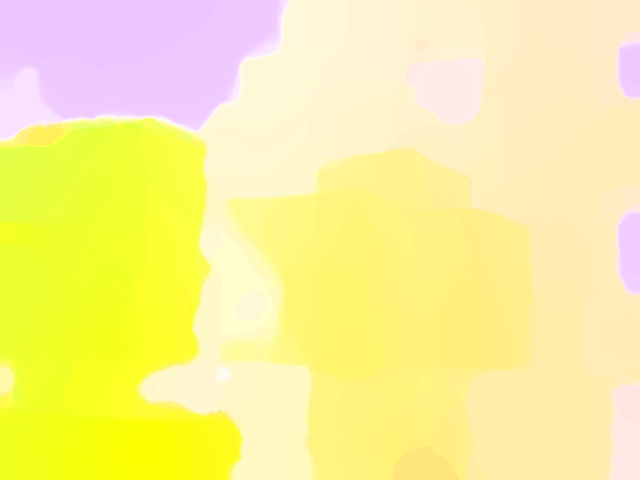} &
 \includegraphics[width=76pt]{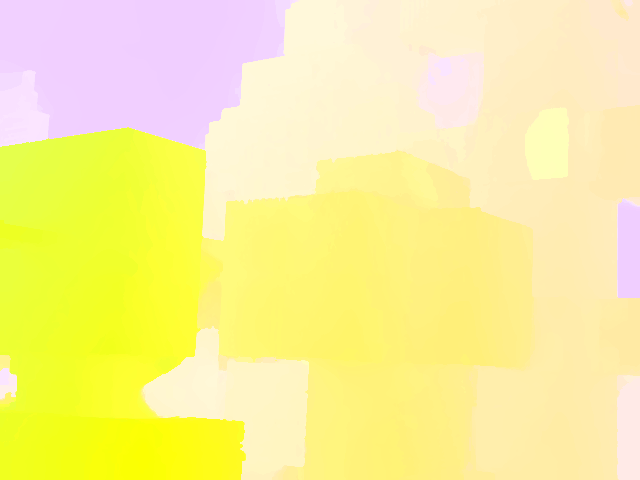} &
  \includegraphics[width=76pt]{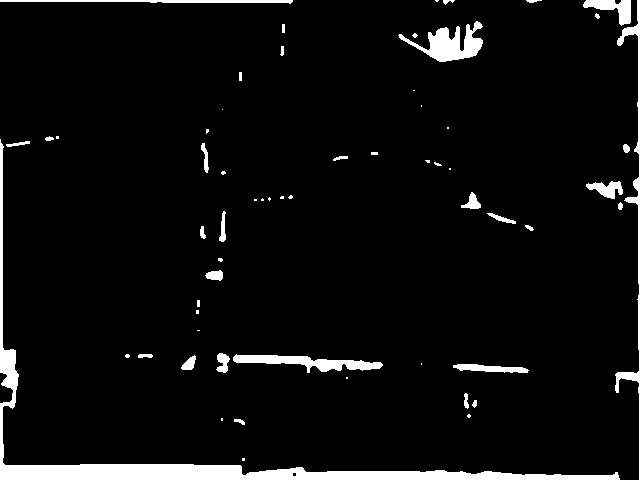} \\
  \mbox{\small \small $I_1$} & \mbox{\small \small Ground truth $\w$}  & \mbox{\small \small AggregFlow $\w$} & \mbox{\small \small DeepFlow \cite{Weinzaepfel13}} & \mbox{\small \small MDP-Flow2 \cite{Xu12}} & \mbox{\small \small AggregFlow
  $o$} \\[5pt]
   \end{array}$

  \caption{Comparative evaluation with \cite{Weinzaepfel13} and \cite{Xu12} on several sequences of the Middlebury dataset. From top to bottom: sequences \textit{grove3, rubberwhale, urban2, urban3}. In each row from left to right: first input image; ground truth motion field; motion field computed resp. with AggregFlow,  DeepFlow \cite{Weinzaepfel13} and MDP-Flow2 \cite{Xu12}; occlusion map computed with AggregFlow.}
\label{fig::visual_results_mid}
  \end{figure}

Recovered occlusion maps are displayed in Fig.\ref{fig::visual_results} and Fig.\ref{fig::visual_results_mid}. For the large occluded regions of
Figure \ref{fig::visual_results} for which ground truth is available, the estimated occlusion map is correct in most cases. A specific behaviour is
particularly prominent in the \textit{market\_5} example, where occlusions are overdetected. It is due to the modeling assumption stating that occluded
regions correspond to large violations of the data constancy equation. Regions of illumination changes may thus be detected as occlusions. While it leads strictly speaking to wrong occlusion detection, it can still be beneficial to motion estimation by implicitly treating illumination changes.

\section{Conclusion}
We have presented a new two-step optical flow estimation method called AggregFlow. It articulates the computation of local motion candidates and their global aggregation while jointly recovering occlusion maps. The framework is generic, and both the local and global steps could be adapted for specific purposes. We demonstrated the added value of combining patch correspondences and patch-based affine motion estimation to produce highly accurate motion candidates, advocating the relevance of patch-based parametric motion estimation, provided size and position of the patches are appropriately defined. The integration of multiple patch correspondences in the candidates generation process allows us to deal with local matching ambiguities. We formulated the aggregation step as a discrete optimization problem, selecting the best motion candidate at every pixel while preserving motion discontinuities and achieving occlusion recovery.
The occlusion scheme acts in both steps of AggregFlow. An exemplar-based occlusion term is incorporated in the global aggregation energy. Incidentally, it could be integrated in other estimation paradigms as well, e.g., in variational approaches. Occlusion cues derived from the computed motion candidates are exploited in the sparse modeling of occlusions. Overall, AggregFlow achieves state-of-the-art results on the MPI Sintel benchmark. The most significant improvements are reached in occluded regions and for large displacements. 

Extensions of the method could tackle remaining matching errors in the patch correspondence and in the exemplar search substeps. A more elaborate and discriminative distance than the pixel-based $L_1$ distance could be envisioned for patch matching. Future work could also deal with a GPU implementation
to largely improve computation efficiency.

\end{document}